%% file: main.tex
\documentclass{article} % For LaTeX2e
\usepackage{colm2024_conference}

\usepackage{microtype}
\usepackage{hyperref}
\usepackage{url}
\usepackage{booktabs}
\usepackage{subcaption}
\usepackage{amsmath}
\usepackage[noblocks]{authblk}
\title{{\sys{}}: \underline{A} \underline{P}rompt \underline{P}rogramming \underline{L}anguage for Harmonious Integration of Programs and Large Language Model Prompts}

\author[1, 2, *]{Honghua Dong}
\author[1, 2, 3, *]{Qidong Su}
\author[1, 2, 3]{Yubo Gao}
\author[1, 2]{Zhaoyu Li}
\author[1, 2]{Yangjun Ruan}
\author[1, 2, 3]{Gennady Pekhimenko}
\author[1, 2]{Chris J. Maddison}
\author[1, 2, 4]{Xujie Si}

\affil{\footnotesize University of Toronto $^2$Vector Institute $^3$CentML $^4$Mila $^*$Equal contribution}
\usepackage{xspace}
\usepackage{makecell}
\usepackage{adjustbox}
\usepackage[frozencache, cachedir=./minted-dir/]{minted}
\usepackage{wrapfig}

\newcommand{\sys}{APPL}
\newcommand{\str}{\texttt{StringFuture}\xspace}
\newcommand{\strs}{\texttt{StringFuture}s\xspace}
\newcommand{\bool}{\texttt{BooleanFuture}\xspace}

\definecolor{mintgreen}{RGB}{98, 153, 5}
\definecolor{mypurple}{HTML}{AD49A2}

\setminted{fontsize=\scriptsize, frame=single}

\usepackage{enumitem}
\setlist[itemize]{leftmargin=*}

\colmfinalcopy % Uncomment for camera-ready version, but NOT for submission.
\isarxiv
\begin{document}

\maketitle

\input{content/abstract}
\input{content/intro}

\input{content/related}
\input{content/language/index}
\input{content/impl}

\input{content/case}

\input{content/conclusion}

\input{content/ack}

% \newpage
\input{content/reproduce}

\bibliography{ref}
\bibliographystyle{colm2024_conference}

\newpage

\appendix
\input{appendix/impl}

\input{appendix/eval}
\input{appendix/trace}

\input{appendix/long_prompt}

\end{document}

%% file: content/abstract.tex
\begin{abstract}
Large Language Models (LLMs) have become increasingly capable of handling diverse tasks with the aid of well-crafted prompts and integration of external tools, but as task complexity rises, the workflow involving LLMs can be complicated and thus challenging to implement and maintain.
To address this challenge, we propose APPL, \underline{A} \underline{P}rompt \underline{P}rogramming \underline{L}anguage that acts as a bridge between computer programs and LLMs, allowing seamless embedding of prompts into Python functions, and vice versa. 
APPL provides an intuitive and Python-native syntax, an efficient parallelized runtime with asynchronous semantics, and a tracing module supporting effective failure diagnosis and replaying without extra costs. 
We demonstrate that APPL programs are intuitive, concise, and efficient through three representative scenarios: Chain-of-Thought with self-consistency (CoT-SC), ReAct tool use agent, and multi-agent chat.
Experiments on three parallelizable workflows further show that APPL can effectively parallelize independent LLM calls, with a significant speedup ratio that almost matches the estimation. \footnote{Project website can be found in \url{https://github.com/appl-team/appl}}

\end{abstract}

%% file: content/intro.tex
\section{Introduction}

Large Language Models (LLMs) have demonstrated remarkable capabilities in understanding and generating texts across a broad spectrum of tasks~\citep{t5,gpt3,palm,llama,llama2,gpt4}. Their success has contributed to an emerging trend of regarding LLMs as novel operating systems~\citep{karpathy2023llmos,ge2023llmao,packer2023memgpt}.
Compared with traditional computer systems that precisely execute structured and well-defined programs written in programming languages, LLMs can be guided through flexible natural language prompts to perform tasks that are beyond the reach of traditional programs.

\input{figures/first-fig}

\input{listing/first_example}

Meanwhile, there is a growing interest in harnessing the combined strengths of LLMs and conventional computing to address more complex challenges.
Approaches like tree-of-thoughts~\citep{yao2024tree}, RAP~\citep{hao2023reasoning} and LATS~\citep{zhou2023language} integrate search-based algorithms with LLM outputs, showing improved outcomes over using LLMs independently.
Similarly, the creation of semi-autonomous agents such as AutoGPT~\citep{richards2023autogpt} and Voyager~\citep{wang2023voyager} demonstrates the potential of LLMs to engage with external tools, like interpreters, file systems, and skill libraries, pushing the boundaries of what integrated systems can achieve.
However, as task complexity rises, the workflow involving LLMs becomes more intricate, requiring more complex prompts guiding LLMs and more sophisticated programs implementing the workflow, which are both challenging to implement and maintain.

To address these challenges, as illustrated in Figure~\ref{fig:vision}, we propose \sys{}, A Prompt Programming Language for harmonious integration of conventional programming and LLMs. More specifically, \sys{}'s key features include:

\textbf{(1) Readability and maintainability via seamless integration with Python.}
As shown in Figure~\ref{fig:first-example}, \sys{} seamlessly embeds natural language prompts to Python programs, maintaining prompts' readability while inheriting modularity, reusability, and dynamism from the host programming language.

\textbf{(2) Automatic parallelization via asynchronous computation.}
\sys{} schedules LLM calls asynchronously, leveraging potential independence among them to facilitate efficient parallelization. This offloads the burden of users to manage synchronization manually, with almost no extra work as shown in Figure~\ref{fig:first-appl}. Its effectiveness is validated in Section~\ref{sec:eval} where experiments show significant accelerations for three popular applications.
\textbf{(3) Smooth transition between structured data and natural language.}
\sys{} enables seamlessly converting program objects into prompts (namely \texttt{promptify}). For example, \texttt{promptify}-ing Python functions converts them into LLM-understandable tool specifications.
In another direction, based on instructor~\citep{instructor}, \sys{} enables the specification of output structures for LLMs, such as ensuring outputs conform to Pydantic's BaseModel~\citep{pydantic}.
When tool specifications are provided, they can be used to constrain the LLM outputs to be a valid tool call.

\sys{} strives to be a powerful, yet intuitive extension of Python and leverages the unique capabilities of LLMs while being compatible with Python's syntax and ecosystem. It empowers developers to easily build sophisticated programs that harness the power of LLMs in a manner that's both efficient and accessible.

%% file: figures/first-fig.tex
\begin{wrapfigure}{r}{0.5\linewidth}
    \vspace{-1em}
    \centering
    \includegraphics[width=\linewidth]{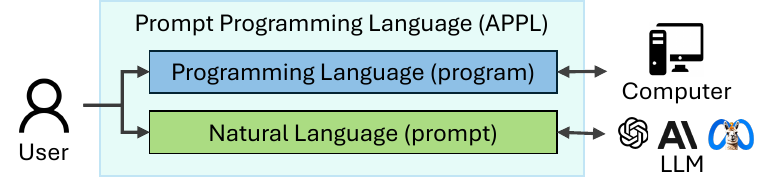}
    \vspace{-1.5em}
    \caption{\small We introduce \sys{}, a \emph{prompt programming language} that integrates conventional programs and natural language prompts to provide a unified interface for users to access computers and LLMs together.\sys{} also facilitates users fusing the strengths of computer programs and LLMs by providing convenient conventions between them.}
    \vspace{-0.5em}
    \label{fig:vision}
\end{wrapfigure}

%% file: listing/first_example.tex
\begin{figure}[tb!]
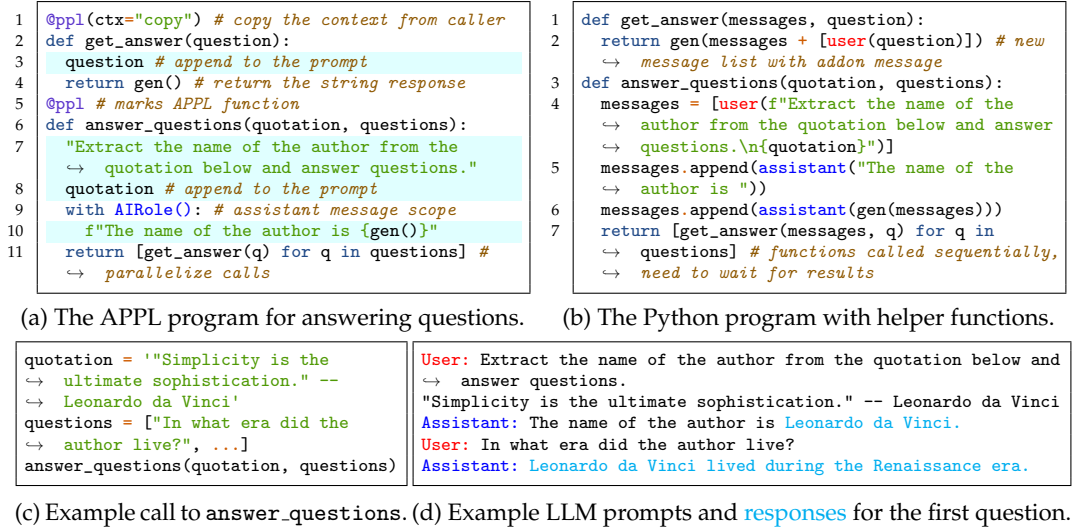


% \vspace{-0.2em}
\begin{subfigure}[c]{0.49\linewidth}
\begin{minted}[numbersep=5pt,breaklines ,linenos,xleftmargin=8pt,highlightlines={3,7,8,10}, escapeinside=||,frame=single]{python}
@ppl(ctx="copy") # copy the context from caller
def get_answer(question):
  question # append to the prompt
  return gen() # return the string response
@ppl # marks APPL function
def answer_questions(quotation, questions):
  "Extract the name of the author from the quotation below and answer questions."
  quotation # append to the prompt
  with |\color{blue}AIRole()|: # assistant message scope
    f"The name of the author is {gen()}"
  return [get_answer(q) for q in questions] # parallelize calls
\end{minted}
\vspace{-0.6em}
\caption{\small The \sys{} program for answering questions.}
\vspace{0.2em}
\label{fig:first-appl}
\end{subfigure}
\hfill
\begin{subfigure}[c]{0.49\linewidth}
\centering
\begin{minted}[numbersep=5pt,breaklines ,linenos,xleftmargin=8pt,highlightlines={},escapeinside=||,frame=single]{python}
def get_answer(messages, question):
  return gen(messages + [|\color{red}user|(question)]) # new message list with addon message
def answer_questions(quotation, questions):
  messages = [|\color{red}user|(f"Extract the name of the author from the quotation below and answer questions.\n{quotation}")]
  messages.append(|\color{blue}assistant|("The name of the author is "))
  messages.append(|\color{blue}assistant|(gen(messages)))
  return [get_answer(messages, q) for q in questions] # functions called sequentially, need to wait for results
\end{minted}
\vspace{-0.6em}
\caption{\small The Python program with helper functions.}
\vspace{0.2em}
\label{fig:first-openai}
\end{subfigure}
% not use separate blocks to display inputs and prompts
% \hfill
\begin{subfigure}[c]{0.373\linewidth}
\centering
\begin{minted}[breaklines,highlightlines={}, frame=single]{python} 
quotation = '"Simplicity is the ultimate sophistication." -- Leonardo da Vinci'
questions = ["In what era did the author live?", ...]
answer_questions(quotation, questions)
\end{minted}
\vspace{-0.5em}
\caption{Example call to \texttt{answer\_questions}.} % Function usage example.
\end{subfigure}
\hfill
\begin{subfigure}[c]{0.627\linewidth}
\centering
\begin{minted}[breaklines,highlightlines={}, frame=single,escapeinside=||]{text} 
|\color{red}User:| Extract the name of the author from the quotation below and answer questions.
"Simplicity is the ultimate sophistication." -- Leonardo da Vinci
|\color{blue}Assistant:| The name of the author is |\color{cyan}Leonardo da Vinci.|
|\color{red}User:| In what era did the author live?
|\color{blue}Assistant:| |\color{cyan}Leonardo da Vinci lived during the Renaissance era.|
\end{minted}
% |\PYG{l+c}{# Generated}|
\vspace{-0.5em}
\caption{Example LLM prompts and {\color{cyan} responses} for the first question.} %  Highlighted lines indicate the completed assistant message.
\label{fig:first-prompt}
\end{subfigure}
\vspace{-0.5em}
\caption{\small \textbf{(a)} The \sys{} program for answering multiple questions about a quotation by first extracting the author's name.
%
% For example, when we ask it: \texttt{In what era did the author live?} with quotation \texttt{"Simplicity is the ultimate sophistication." -- Leonardo da Vinci}, its answer would be like \texttt{Renaissance Era}.
%
In this \sys{} program, the \texttt{@ppl} marks the function to be \sys{} function, which provides a \emph{context} that interactive expression statements (highlighted lines) can be interacted with.
Passing the context across \sys{} functions is implicitly handled along with function calls, mimicking the process of stack frame management.
The \texttt{gen} function requests an LLM generation using the prompts stored in the context, with other optional arguments following the OpenAI's API~\citep{openaiapi}. For instance, the \texttt{gen} in line 4 uses the prompt composed by (highlighted) lines 7,8,10 (since context is copied), and 3.
The LLM generation is computed asynchronously and only synchronizes when the value is needed, therefore independent calls are automatically parallelized as shown in line 11. 
\textbf{(b)} A Python program implementing the same function uses more codes, even with the helper functions. Furthermore, extra codes are needed to parallelize \texttt{get\_answer} function calls.
\textbf{(c)} Example usage for the program and \textbf{(d)} its corresponding LLM prompts and example responses.
}
\label{fig:first-example}
\vspace{-0.5em}
\end{figure}

%% file: content/related.tex
\section{Related Work}
\label{sec: related}

\input{tables/cmp}

\textbf{LLM programs.}
The programs involving LLMs continue evolving sophisticated to tackle more challenging tasks, from few-shot prompts~\citep{gpt3} to elaborated thought process~\citep{wei2022cot,ning2023skeleton,yao2024tree,got}, from chatbot~\citep{chatgpt} to tool-use agents~\citep{nakano2021webgpt,ahn2022saycan,liu2022mind,liang2023code,richards2023autogpt,patil2023gorilla,wang2023voyager,qin2023toolllm}, virtual emulators~\citep{ruan2024toolemu}, operating systems~\citep{packer2023memgpt}, and multi-agent systems~\citep{park2023genagents,hong2023metagpt,qian2023communicative}. The increasing complexity necessitates powerful frameworks to facilitate the development of those programs.

\textbf{LLM software stack.}
On the low level, LLM services are provided by high-performance backends~\citep{llamacpp,tgi,vllm} as APIs like OpenAI APIs~\citep{openaiapi}.
Building on this, modules with specific functionality are invented, like unifying APIs~\citep{litellm}, constraining LLM outputs~\citep{outlines,instructor} and parallelizing function calls~\citep{llmcompiler}, where \sys{} utilizes some of them as building blocks. 
LMQL~\citep{lmql}, Guidance~\citep{guidance},
and SGLang~\citep{sglang} are languages similar to \sys{}. These languages provide developers with precise tools for crafting sophisticated model interactions, enabling finer-grained control over LLM behaviors. We compare \sys{} with them by case studies in Section~\ref{sec:case} and summarize the comparison in Table~\ref{tab:related}, where \sys{} provides more automatic supports for prompting and parallelizing LLM calls to streamline the whole workflow.
There are also higher-level general-purpose frameworks~\citep{askit,marvin,dspy,langchain}
and agent-specific frameworks~\citep{xagent2023,li2023camel,wu2023autogen}. The pipelines in these frameworks can also be implemented using prompt languages like \sys{}.

% A comparison among LMQL, Guidance, and SGLang is presented in Table 1

% # high-level frameworks
% ## agent-specific frameworks
% - xagent~\citep{xagent2023}
% - camel~\citep{li2023camel}
% - autogen~\citep{wu2023autogen} % multiagent

% ## general-purpose frameworks
% - LangChain~\citep{langchain}
% - DSPy~\citep{dspy}
% - marvin~\citep{marvin}
% - AskIt~\citep{askit}

% # Domain-specific languages (DSLs)
% - LMQL
% - Guidance
% - SGLang

% # module for a specific functionality
% - LLMCompiler~\citep{llmcompiler}
% - litellm~\citep{litellm}
% - outlines~\citep{outlines}
% - instructor~\citep{instructor}

% # local models serving
% - vllm~\citep{vllm}
% - llamacpp~\citep{llamacpp}
% - TGI~\citep{tgi}.

%% file: tables/cmp.tex
\begin{table}[tb]
\footnotesize
    \centering
    \vspace{-0.5em}
    \begin{adjustbox}{width=\linewidth}
    \begin{tabular}{l | c | c | c | c}
    \toprule
     Features & \makecell{LMQL\\\citep{lmql}} & \makecell{Guidance\\\citep{guidance}} & \makecell{SGLang\\\citep{sglang}} & \textbf{\sys{}} \\
    \midrule
    Language Syntax & Python-like & Python & Python & Python\\
    Parallelization and Asynchrony & Sub-optimal & Manual & \textbf{Auto} & \textbf{Auto} \\
    \midrule
    Prompt Capturing & \textbf{Auto} & Manual & Manual & \textbf{Auto} \\
    Prompt Context Passing & Limited & Manual & Manual & \textbf{Auto}\\
    Promptify Tool & Manual & Manual & Manual & \textbf{Auto} \\
    Prompt Compositor & No & No & No & \textbf{Yes} \\
    \midrule
    Generation Output Retrieval & Template variables & \texttt{str}-index & \texttt{str}-index & \textbf{Python-native} \\
    Generation without Side-effects & No & No & No & \textbf{Yes} \\
    \midrule
    Failure Recovery & No & No & No & \textbf{Yes} \\

    %Python Control Flow & Yes & Yes & Interpreter only & Python native \\
    \bottomrule
    \end{tabular}
    \end{adjustbox}
    \vspace{-0.5em}
    \caption{\small Comparison across Prompt Languages.
    \sys{} enhances the user experience of crafting LLM-augmented programs with Python-native syntax while achieving high performance with an efficient parallelizable runtime system.
    We carefully design the language features involved in prompting and generation, which streamlines the whole workflow.
    \sys{} also provides failure recovery through its tracing mechanism.
    See Sections~\ref{sec:lang} for the language design and Section~\ref{sec:case} for more comparison details.
    }
    \label{tab:related}
    \vspace{-1.5em}
\end{table}

    % \begin{adjustbox}{width=\linewidth}
    % \begin{tabular}{c | c |  c | c | c | c | c | c | c}
    % \toprule
    % & Parallelize & \makecell{Context\\Manage} & 
    % \makecell{Output\\Retrieval} &
    % \makecell{Python\\Control\\Flow}  &  \makecell{Prompt\\Compositors}  & 
    % \makecell{OpenAI\\Tool\\Calling} & Tracing
    % \\
    % \midrule
    % LMQL & Sub-optimal &  Auto  & Template variables & Yes  & No & No & No\\
    % Guidance & Manual  & Manual & str-index & Yes  & No  & No & No\\
    % SGLang & Auto  & Manual & str-index & Interpreter only  & No  & No & No\\
    % \midrule
    % \sys & Auto & Auto & Yes & Python native & Yes & Yes & Yes\\

    % \bottomrule
    % \end{tabular}
    % \end{adjustbox}

%% file: content/language/index.tex
\section{Language Design and Features}
\label{sec:lang}

\subsection{Design Principles of \sys{}}

% NL into code
\textbf{(1) Readability and Flexibility.}
Natural language provides a readable and understandable human-LLM interface, while programming languages enable flexible controls, clear structures, modularization, reusability, and external tools. \sys{} aims to utilize both strengths.

\textbf{(2) Easy Parallelization.}
% LLMs can be regarded as a different computational resource from conventional computing, its asynchronous nature can be utilized for easy parallelization.
% LLMs and other computational resources form a heterogeneous computational system, of which the asynchronous nature can be utilized for easy parallelization.
Parallelization is a critical performance optimization for LLM text generation~\citep{orca,vllm}.
%In LLM calls without dependencies. 
\sys{} aims to provide transparent support for parallelization with almost no extra code introduced.

% code object <-> NL
\textbf{(3) Convenient Conversion.} 
\sys{} aims to facilitate easy conversion between structured data (including codes) used in programming and the natural language used in LLMs.
% The language should provide convenient ways to transform program objects into prompts (promptify, similar to serialization), and from LLM generation results to program objects (parse).

\input{figures/runtime}

\input{content/language/syntax}

\input{content/language/runtime}

%% file: figures/runtime.tex
\begin{figure}[tb]
    \centering
    \vspace{-0.5em}
    \includegraphics[width=\linewidth]{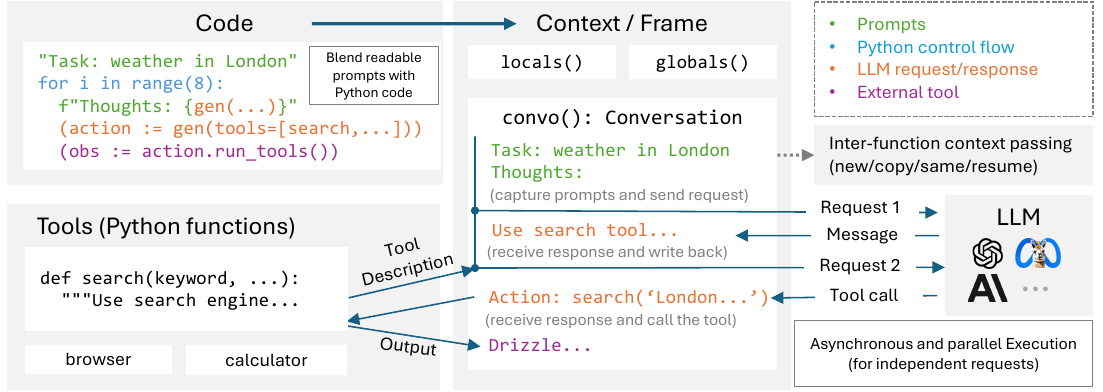}
    \vspace{-1.5em}
    \caption{\small Overview of the \sys{} runtime. The code is executed following \textcolor{cyan}{Python control flows}, with \textcolor{mintgreen}{prompts} and \textcolor{orange}{LLM responses} being captured into the \texttt{conversation} (can be retrieved by \texttt{convo()}, similar to \texttt{locals()} and \texttt{globals()} in the Python frame). The \textcolor{orange}{\texttt{gen}} uses the \texttt{conversation} captured so far as prompts to call LLMs and is executed asynchronously, so that independent calls can be parallelized easily.
    \sys{} allows using \textcolor{mypurple}{Python functions as tools} for LLMs by extracting information from their docstrings and signatures. When tools are provided, their specifications are included in the request and the returned tool calls can be easily \textcolor{mypurple}{executed to get results}. See detailed code in Figure~\ref{fig:react-appl}.
    }
    \label{fig:runtime}
    \vspace{-1.5em}
\end{figure}

%% file: content/language/syntax.tex
\subsection{\sys{} Syntax and Semantics}
\label{sec:syntax}

As illustrated in the \sys{} program of Figure~\ref{fig:first-appl}, the syntax and semantics of \sys{} are extended from Python with slight modifications to integrate natural language prompts and LLMs calls into the traditional programming constructs deeply.

\textbf{\sys{} functions} are the fundamental building blocks of \sys{}, denoted by the \texttt{@ppl} decorator.
Each \sys{} function has an implicit context to store prompt-related information. Within \sys{} functions, the expression statements are interpreted as ``interactions'' with the context,
where the typical behavior is to append strings to the prompt.

\textbf{Expression statements} are standalone Python statements formed by an expression, for example, \texttt{"str"}, while an assignment (e.g., \texttt{a = "str"}) is not. Their evaluation result is handled differently according to the environment. For example, it causes no extra effect in standard Python while the result is displayed in the interactive environment in Jupyter. Similarly, \sys{} provides an environment where each \sys{} function contains a scratchpad of prompts (i.e. context), and the values of expression statements will be converted to prompts (via \texttt{promptify}) and appended to the prompt scratchpad.
For example, a standalone string as line 7 or 8 in Figure~\ref{fig:first-appl} will be appended to the prompt scratchpad.

\textbf{LLM generations} (\texttt{\color{blue}gen}), when being called, implicitly use the accumulated prompt in the context of the \sys{} function. A special case is the standalone f-string, which is split into parts and they are captured as prompts in order, detailed in Appendix~\ref{app:fstr}. This provides a more readable way to combine prompts and LLM generations in one line. For example, line 10 of Figure~\ref{fig:first-appl} means start generation by first adding the prefix ``The name of the author is '' to the prompt, which matches human intuition.

\input{listing/comp_def}

\textbf{Context managers} modify how prompt statements are accumulated to the prompt in the context. \sys{} provides two types of context managers: 
\\
\textbf{(1)} \emph{Role Changers} (such as \texttt{AIRole} in Figure~\ref{fig:first-appl}) are used in chats to specify the owner of the message, and can only be used as the outmost one.
\\
\textbf{(2)} \emph{Prompt Compositors} (such as \texttt{NumberedList} in Figure~\ref{fig:comp_def}) specify the delimiter, indention, indexing format, prolog, and epilog. These compositors allow for meticulously structured finer-grained prompts in hierarchical formatting through native Python syntax, greatly enhancing flexibility without losing readability.
% Neither LMQL~\citep{lmql} nor Guidance~\citep{guidance} provides such utilities for format control.
% See more usage examples in Appendix\ref{app:long-prompt}.

\textbf{Definition} is a special base class for custom definitions. As shown in Figure~\ref{fig:comp_def}, once declared, \emph{Definitions} can be easily referred to in the f-string, or used to create an instance by completing its description. They are useful for representing concepts that occur frequently in the prompt and support varying the instantiation in different prompts. See more usage in Appendix~\ref{app:long-prompt} when building a long prompt.

%% file: listing/comp_def.tex
% numbersep=5pt,breaklines ,linenos,xleftmargin=10pt,

\begin{wrapfigure}{r}{0.45\linewidth}
\centering
\vspace{-1.2em}
\begin{subfigure}[c]{\linewidth}
\begin{minted}
[highlightlines={1,7,8,9}, frame=single]
{python}
Req = define("Requirement")
InputReq = define(f"Input {Req}")
OutputReq = define(f"Output {Req}")
@ppl
def func():
  f"... the following {Req}s:"
  with NumberedList(indent=2):
    InputReq(desc="Input should ...")
    OutputReq(desc="Output should ...")
  return records()
\end{minted}
\end{subfigure}
\\
\begin{subfigure}[c]{\linewidth}
\begin{minted}[frame=single]{bash}
>>> print(func())
... the following Requirements:
  1. Input Requirement: Input should ...
  2. Output Requirement: Output should ...
\end{minted}
\end{subfigure}
\caption{\small Illustration of the usage of \emph{Definitions} and \emph{Prompt Compositors}.}
\label{fig:comp_def}
\vspace{-2em}
\end{wrapfigure}

%% file: content/language/runtime.tex
\subsection{\sys{} Runtime}\label{sec: runtime}

\textbf{\sys{} runtime context.} Similar to Python's runtime context that contains local and global variables (\texttt{locals()} and \texttt{globals()}), the context of \sys{} functions contains local and global prompts that can be accessed via \texttt{records()} and \texttt{convo()}. The example in Figure~\ref{fig:runtime} illustrates how the prompts in the context changed during the execution.

\input{figures/ctx_methods}

\textbf{Context passing across functions.}
When a \sys{} function (caller) calls another \sys{} function (callee), users might expect different behaviors for initializing the callee's context depending on specific use cases. 
\sys{} provides four ways as shown in Figure~\ref{fig:ctxpass}: \\
\textbf{(1)} \texttt{new}: Creates a clean context. This is the default and most common way. \\
\textbf{(2)} \texttt{copy}: Creates a copy of the caller's context. This is similar to ``call by value'' where the modification to the copy will not influence the original context. This is useful for creating independent and asynchronize LLM calls as shown in Figure~\ref{fig:first-appl}. \\
\textbf{(3)} \texttt{same}: Uses the same context as the caller. This is similar to ``call by reference'' where the modifications are directly applied to the original context. \\
\textbf{(4)} \texttt{resume}: Resumes callee's context from the last run. This makes the function stateful and suitable for building interactive agents with history (see Figure~\ref{fig:agent-b} for an example).
% ! This will make the function stateful, use it with caution.

When decomposing a long prompt into smaller modules, both \texttt{new} and \texttt{same} can be used as shown in Figure~\ref{fig: ctxpass-example}, but the default way (\texttt{new}) should be prioritized when possible.

\textbf{Asynchronous semantics.}
To automatically leverage potential independence among LLM calls and achieve efficient parallelization, we borrow the idea of the asynchronous semantics from PyTorch~\citep{pytorch}.
% where the main thread does not immediately wait for the outcome of a GPU operator, but only synchronizes when necessary, such as printing, copying the data across devices, or when involving control flow.
Asynchronous semantics means the system can initiate operations that run independently of the main execution thread, allowing for parallelism and often improving performance.
In \sys{}, LLM calls (\texttt{gen}) are asynchronous, which means the main thread proceeds without being blocked by \texttt{gen} until accessing the generation results is necessary. See implementation details in Section~\ref{sec: strfuture}.
%
% However, manually managing synchronization imposes a heavy burden on users, especially for developers unfamiliar with asynchronous programming.
% To leverage potential parallel chance, we propose a new data structure called \emph{\str{}}. 
% Inspired by common practice in asynchronous programming~\citep{future}, 
%\str{} represents a generation request sent to the LLM backend, which may still be in progress. 
% When a \str{} object is instantiated, \sys{} starts a backend thread to handle the request, ensuring that the main thread proceeds without being blocked. 
% The detailed description of its implementation is in Section~\ref{sec: strfuture}.

\input{listing/tool_call}

\textbf{Tool calling.}
Tool calling is an indispensable ability for LLM agents to interact with external resources. To streamline the process of providing tools to LLMs, \sys{} automatically creates tool specifications from Python functions by analyzing their docstrings and signatures (detailed in Appendix~\ref{app:tool_spec}). As shown in Figure~\ref{fig:toolcall}, the Python functions \texttt{is\_lucky} and \texttt{search} can be directly utilized as tools for LLMs. Therefore, thanks to the well-written docstrings, the functions from Python's rich ecosystem can be integrated into LLMs as tools with minimal effort.
Moreover, \sys{} automatically converts LLMs' outputs as function calls that can be directly executed to obtain results.

\textbf{Tracing and caching.}
\sys{} supports tracing \sys{} functions and LLM calls to facilitate users to understand and debug the program executions.
The trace is useful for reproducing (potentially partial) execution results by loading cached responses of the LLM calls, which enables failure recovery and avoids the extra costs of resending these calls. 
This also unlocks the possibility of debugging one LLM call conveniently. Users can also visualize the program execution to analyze the logic, expense, and runtime of the program.
\sys{} provides two modes of tracing, namely strict and non-strict modes, corresponding to different reproduction requirements (perfect or statistical reproduction).
We further explain the implementation details for handling the asynchronous semantics in Appendix~\ref{app:trace-impl}, and demonstrate the trace visualization in Appendix~\ref{app:trace}.

%% file: figures/ctx_methods.tex
\begin{figure}[tb]
% \vspace{-1em}
\centering
\begin{adjustbox}{minipage=\linewidth,scale=0.85}
\begin{subfigure}[b]{0.25\linewidth}
\includegraphics[width=\linewidth]{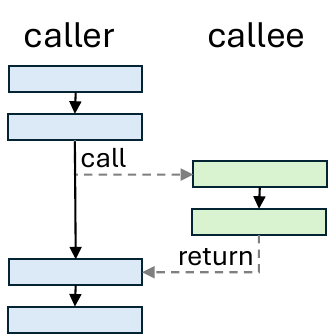}
\caption{new}
\end{subfigure}
\hfill
\begin{subfigure}[b]{0.25\linewidth}
\includegraphics[width=\linewidth]{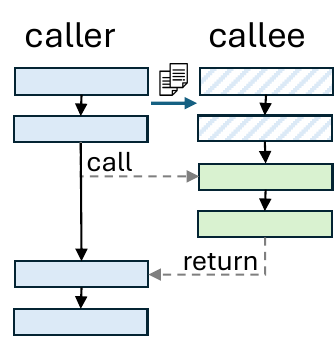}
\caption{copy}
\end{subfigure}
\hfill
\begin{subfigure}[b]{0.225\linewidth}
\includegraphics[width=\linewidth]{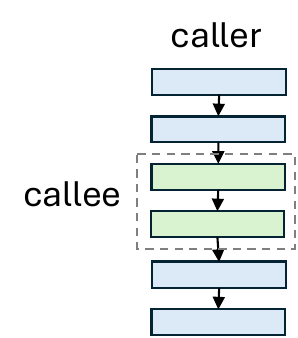}
\caption{same}
\end{subfigure}
\hfill
\begin{subfigure}[b]{0.22\linewidth}
\includegraphics[width=\linewidth]{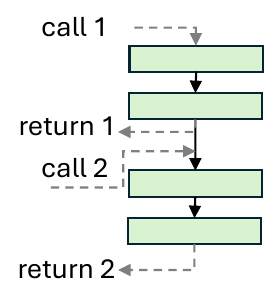}
\caption{resume}
\end{subfigure}
\end{adjustbox}
\vspace{-0.5em}
\caption{\small Illustration of four different ways of inter-function context passing. Usage examples can be found in Section~\ref{sec:multi-agent} and Appendix~\ref{app:ctx}. Prompts connected by solid arrows belong to the same prompt context. Dashed arrows represent the control flow of function calls and returns.}
\label{fig:ctxpass}
\vspace{-1em}
\end{figure}

%% file: listing/tool_call.tex
\begin{wrapfigure}{r}{0.43\linewidth}
\vspace{-2em}
\begin{minted}[numbersep=5pt,breaklines ,linenos,xleftmargin=8pt,frame=single,highlightlines={9,11}]{python}
def is_lucky(x: int) -> bool:
  """Determine whether the input number is a lucky number.
  <Args and Returns omitted ... >"""
  return sympy.isprime(x+3)
@ppl
def answer_with_tool(question: str):
  question # E.g., Is 2024 a lucky number?
  # Generation with tool integration
  action = gen(tools=[is_lucky, search])
  # Execute the tools and get the results
  results = action.run_tool_calls() 
  return results[0].content
\end{minted}
% def search(query: str) -> str: ...
\vspace{-1.1em}
\caption{\small Tool calling example in \sys{}.}
\label{fig:toolcall}
\vspace{-2em}
\end{wrapfigure}

%% file: content/impl.tex
\section{Implementation}

We elaborate on the implementation details of \sys{}, including compilation and asynchronous execution. 

\subsection{Compilation}

To provide the runtime context for \sys{} functions more smoothly and make it transparent to users, we choose to compile (more specifically, transpile) the Python source code to insert the context into the code.
We use Python's \texttt{ast} package to parse the original function as AST, mainly apply the following code transformations on the AST, and compile back from the modified AST to a Python function.

\begin{itemize}
% \vspace{-0.5em}
    \item \textbf{Context management.} We use the keyword \texttt{\_ctx} to represent the context of the function, which is inserted in both the definition of \sys{} functions and the function calls that need the context.
    The functions that require the context as inputs are annotated by the attribute \texttt{\_\_need\_ctx\_\_}, including all functions introduced in Section~\ref{sec:syntax}: \sys{} functions, \texttt{gen} and context manager functions.
    \item \textbf{Capture expression statements.} We add an execution function as a wrapper for all expression statements within the \sys{} function and let them ``interact'' with the context. The execution behavior differs based on the value types, where the behavior for the most common \texttt{str} type is appending it to the prompt of the context. For a standalone f-string, we split it into several statements to let them be ``executed'' in order so that \texttt{gen} functions in the f-string can use the text before them. This procedure is detailed in Appendix~\ref{app:fstr}.
% \vspace{-0.5em}
\end{itemize}

\subsection{Asynchronous Execution and Future Objects}
\label{sec: strfuture}

To implement the asynchronous semantics, we adopt the concept of \texttt{Future} introduced in the Python standard library \texttt{concurrent.futures}, which represents the asynchronous execution of a callable. When the result of \texttt{Future} is accessed, it will wait for the asynchronous execution to finish and synchronize the result.

We introduce \str{}, which represents a string object, but its content may not be ready yet. \str{}s behave almost identically to the normal \texttt{str} objects in Python, so that users can use them as normal strings transparently and enjoy the benefits of asynchronous execution. Meanwhile, the \str{}  delays its synchronization when possible, ideally only synchronizing when the \texttt{str} method is called.

Formally, we define \str{} $S$ to be a list of \strs{}: $[s_1, s_2, \dots, s_n]$.
Without loss of generality, we can regard a normal string as an already synchronized \str{}. A \str{} $S$ is synchronized when \texttt{str}($S$) is called, and it collapses to a normal string by waiting and concatenating the results of \texttt{str}($s_i$):
$$\texttt{str}([s_1, \dots, s_n]) \rightarrow {\texttt{str}(s_1)}\oplus \texttt{str}([s_2, \dots, s_n]),$$
where $\oplus$ operation means (string) concatenation.
Such representation enables delaying the concatenation ($\oplus$) of two \strs{} $S$ and $T$ as concatenating ($\oplus$) two lists:
$$S \oplus T = [s_1, \dots, s_n] \oplus [t_1, \dots, t_m] = [s_1, \dots, s_n, t_1, \dots, t_m].$$
For other methods that cannot be delayed, like \texttt{len}($S$), it falls back to materializing $S$ to a normal string and then calls the counterpart method.

Similarly, we introduce \bool{} to represent a boolean value that may not be synchronized yet, which can be used to represent the comparison result between two \strs{}. A \bool{} $B$ is synchronized only when \texttt{bool(B)} is called, which enables using future objects in control flows.

%% file: content/case.tex
% Our advantages
% 1. Auto parallel (StringFuture)
% 2. easy tool call
% 3. structured prompt
\section{Case Studies and Language Comparisons}
\label{sec:case}

In this section, we compare \sys{} with other prompt languages including LMQL ~\citep{lmql}, Guidance ~\citep{guidance}, and SGLang ~\citep{sglang} (whose syntax is mostly derived from Guidance) through usage scenarios, including parallelized LLM calls like the chain of thoughts with self-consistency (CoT-SC) ~\citep{wang2022self}, tool use agent like ReAct~\citep{yao2023react}, and multi-agent chat. We further include a re-implementation of a long prompt (more than 1000 tokens) from ToolEmu~\citep{ruan2024toolemu} using \sys{} in Appendix~\ref{app:long-prompt}, demonstrating the usage of \sys{} functions and prompt compositors for modularizing and structuring long prompts.

\subsection{Chain of Thoughts with Self-Consistency (CoT-SC)}\label{sec:case-cot}

\input{listing/cot_cmp}

As shown in Figure~\ref{fig:cot-compare}, 
when implementing CoT-SC, \sys{} exhibits its conciseness and clarity compared with LMQL, SGLang, and Guidance. We mainly compare them from the following aspects:

\textbf{Parallization.}
% APPL
\sys{} performs automatic parallelization while being transparent to the programmer. In particular, since the returned value need not be immediately materialized, the different branches can execute concurrently.
% SGLang
SGLang allows for parallelization between subqueries by explicitly calling the \texttt{lm.fork()} function.
%LMQL
LMQL employs a conservative parallelization strategy, which waits for an \texttt{async} function immediately after it's called. To support the same parallelization strategy, we use the \texttt{asyncio} library to manually launch LMQL queries in parallel at the top level. 
% Guidance
Guidance does not include asynchrony or parallelization as a core feature. 

\textbf{Generation and output retrieval.}
% Gen
For LLM generation calls, \sys{}, SGLang, and Guidance use a Python function while LMQL uses inline variables (with annotations), which is less flexible for custom generation arguments.
% retrieve
For the results of generations, both SGLang and Guidance store them as key-value pairs in a dictionary associated with the context object and are indexed with the assigned key when required.
LMQL accesses the result by the declared inline variable, but the variable is not Python-native.
In \sys{}, \texttt{gen} is an independent function call, whose return value can be assigned to a Python variable
%(via the walrus operator ``\texttt{:=}'')
, which is \emph{Python-native} and allows for better interoperability with IDEs.
% APPL: separate reading and writing
Moreover, \sys{} separates the prompt reading and writing for the generation. Users have the option to write generation results back to the prompt or not  (called \emph{without side-effects} in Table~\ref{tab:related}). This allows a simple replication of LLM generation calls sharing the same prompt. Otherwise, explicit copying or rebuilding of the prompt would be necessary for such requests.

\textbf{Context management and prompt capturing.} %
% Guidance and SGLang
Both SGLang and Guidance need to explicitly manage context objects (\texttt{s} and \texttt{lm}) as highlighted as {\color{red} red} in Figure~\ref{fig:cot-compare}. The prompts are captured manually using the \texttt{+=} operation. The branching of the context is implemented differently, where SGLang uses the explicit \texttt{fork} method and Guidance returns a new context object in the overridden \texttt{+=} operation.
% APPL & LMQL
In contrast, \sys{} automatically captures standalone expression statements into the prompt context, similar to LMQL which captures standalone strings. The context management is also done automatically for nested function calls in \sys{} and LMQL, where \sys{} four patterns \emph{new}, \emph{copy}, \emph{same}, and \emph{resume} are supported in \sys{} (see Figures~\ref{fig:ctxpass} and \ref{fig:first-appl}) while LMQL only supports \emph{new} and \emph{copy}.

\subsection{Parallelized LLM Calls}
\label{sec:eval}

We validated the effectiveness of parallelizing LLM calls in \sys{} in three tasks: question answering with CoT-SC, content generation with skeleton-of-thought (SoT)~\citep{ning2023skeleton}, and hierarchical summarization with MemWalker~\citep{memwalker}.
As shown in Table~\ref{tab:eval}, the parallel version achieves significant speedup across different branching patterns, and the actual speedup ratios almost match the estimated ones.
More experimental details can be found in Appendix~\ref{app:eval}.

\input{tables/eval_cmp}

\subsection{ReAct Tool Use Agent}
\label{sec:case:react}

We use the ReAct~\citep{yao2023react} tool-use agent as an example to compare the accessibility of tool calls in different languages.
As shown in Figure~\ref{fig:react-appl}, the major workflow of ReAct unfolds over several cycles, each comprising three stages: \textbf{(1)} generating ``thoughts'' derived from existing information to steer subsequent actions; \textbf{(2)} selecting appropriate actions (tool calls) to execute; \textbf{(3)} retrieving the results of executing the actions. These results then serve as observations and influence the subsequent cycle.

\input{listing/react_example}

LLM Tool calling can be divided into three steps: 1) encoding the external tools as specifications that are understandable by LLMs, 2) parsing generated tool calls, and 3) executing the tool calls.
\sys{} provides out-of-the-box support for these steps based on the OpenAI's API~\citep{openaiapi}. As shown in Figure~\ref{fig:react-appl}, documented Python functions like \texttt{is\_lucky} defined in Figure~\ref{fig:toolcall} can directly be used as the tools argument without manual efforts.
Differently, in Figure~\ref{fig:react-guidance}, Guidance (similarly for SGLang and LMQL) can generate tool calls and parse structured arguments via its \texttt{select} and other constraint primitives, but the tool specifications and format constraints need to be written manually.
Guidance provides another way to automate this process but with limited support that only works for functions with string-typed arguments.

\subsection{Complex context management: Multi-agent chat bot}
\label{sec:multi-agent}
\input{listing/multiagent}

We consider a simplified yet illustrative scenario of agent communities~\citep {park2023genagents,hong2023metagpt,qian2023communicative}, where two agents chat with each other, as demonstrated in Figure~\ref{fig:agent-demo}.
During the chat, each agent receives messages from the counterpart, maintains their own chat history, and makes responses.
\sys{} provides flexible context-passing methods for implementing such agents and we illustrate two ways in Figure~\ref{fig:agent} as examples.

\textbf{(a) Explicit chat history:} As shown in Figure~\ref{fig:agent-a}, the conversation history is explicitly stored in \texttt{self.\_history}. The history is first initialized in the \texttt{\_setup} function. In the \texttt{chat} function, the history is first retrieved and written into a new prompt context. At the end, \texttt{self.\_history} is refreshed with \texttt{records()} that represents updated conversation with new messages and responses.

\textbf{(b) \texttt{resume} the context:} As shown in Figure~\ref{fig:agent-b}, each call to the \texttt{chat} method will resume its memorized context, allowing continually appending the conversation to the context. The context is initialization in the first call by copying the caller's context.
This approach provides a clear, streamlined, concise way to manage its own context.

\subsection{Code Succinctness Comparison}\label{sec:code-succ}

As demonstrated in Figure~\ref{fig:cot-compare}, \sys{} is more concise than other compared languages to implement the Cot-SC algorithm. Quantitatively, we further introduce a new metric \texttt{AST-size} to indicate the succinctness of programs written in different languages, which counts the number of nodes in the abstract syntax tree (AST) of the code. For Python programs, we can obtain their AST 
using the standard library with \texttt{ast.parse}.

\input{tables/code_ast_num}

As shown in Table~\ref{tab:code-ast}, LMQL, SGLang, and Guidance need about twice AST nodes than \sys{} for implementing the same tasks, highlighting the succinctness of \sys{}.

%% file: listing/cot_cmp.tex
\begin{figure}[tb]
% \vspace{-2.5em}
\begin{minipage}[tb]{0.49\linewidth}

\begin{subfigure}[tb]{\linewidth}

\begin{minted}[numbersep=5pt,breaklines,linenos,xleftmargin=8pt,escapeinside=||]{python}
@ppl
def cot(num_trials: int):
  "... (three examples omitted)"
  "Q: You need to cook 9 eggs."
  "A: Let’s think step by step."
  return [|\color{cyan}gen()| |\PYG{l+c}{\color{blue}\PYGZsh{} parallel}|
    for _ in range(num_trials)]

answers = cot(num_trials)
\end{minted}
\vspace{-0.6em}
\caption{\small APPL}
\label{fig:cot-appl}
\vspace{0.3em}
\end{subfigure}

\begin{subfigure}[tb]{\linewidth}
% use |\PYG{l+s}{ content }| to escape inside a string, use \PYGZdq{} for double quote and \PYGZdq{} for single quote.
\begin{minted}[numbersep=5pt,breaklines,linenos,xleftmargin=8pt,escapeinside=||]{python}
@lmql.query
async def cot():
  |\PYG{l+c}{\PYGZsq{}\PYGZsq{}\PYGZsq{}lmql}|
  "... (three examples omitted)"
  "Q: You need to cook 9 eggs."
  |\PYG{l+s}{\PYGZdq{}A: Let’s think step by step.{\color{cyan}[ANSWER]}\PYGZdq{}}|
  return |\color{cyan}ANSWER|
  |\PYG{l+c}{\PYGZsq{}\PYGZsq{}\PYGZsq{}}|

async def cot_parallel(num_trials: int):
  responses = [cot() for i in range(num_trials)]
  return |\color{blue}await asyncio.gather(*responses)|

answers = |\color{blue}asyncio.run(cot\_parallel)|
\end{minted}
\vspace{-0.6em}
\caption{\small LMQL}
\label{fig:cot-lmql}
\end{subfigure}

\end{minipage}
\hfill
\begin{minipage}[tb]{0.49\linewidth}

\begin{subfigure}[tb]{\linewidth}
\begin{minted}[numbersep=5pt,breaklines,linenos,xleftmargin=8pt,escapeinside=||]{python}
@function
def cot(|\color{red}s|, num_trials: int):
  |\color{red}s +=| "... (three examples omitted)\n"\
       "Q: You need to cook 9 eggs.\n"\
       "A: Let's think step by step."
  |\color{red}forks = s.fork(num\_trials)|
  for fork in forks:
    |\color{red}fork +=| |\color{cyan}gen(\PYGZdq{}answer\PYGZdq{})| |\PYG{l+c}{\color{blue}\PYGZsh{} parallel in forks}|
  return [|\color{cyan}fork[\PYGZdq{}answer\PYGZdq{}]| for fork in forks]

answers = cot.run(num_trials).ret_value
\end{minted}
\vspace{-0.6em}
\caption{\small SGLang}
\label{fig:cot-sgl}
\vspace{0.3em}
\end{subfigure}

\begin{subfigure}[tb]{\linewidth}
% numbersep=5pt,linenos,xleftmargin=8pt
\begin{minted}[numbersep=5pt,breaklines,linenos,xleftmargin=8pt,escapeinside=||,highlightlines={}]{python}
def cot(|\color{red}lm|, num_trials: int):
  |\color{red}lm +=| "... (three examples omitted)\n"\
        "Q: You need to cook 9 eggs.\n"\
        "A: Let's think step by step."
  answers = []
  for i in range(num_trials): |\PYG{l+c}{\color{blue}\PYGZsh{} sequential}|
    |\color{red}fork = lm| # avoid mutating lm 
    |\color{red}fork +=| |\color{cyan}gen(\PYGZdq{}answer\PYGZdq{})| # become a new object
    answers.append(|\color{cyan}fork[\PYGZdq{}answer\PYGZdq{}]|)
  return answers

answers = cot(lm, num_trials)
\end{minted}
\vspace{-0.6em}
\caption{\small Guidance}
\label{fig:cot-guidance}
\end{subfigure}

\end{minipage}
\vspace{-0.3em}
\caption{
\small Comparison of CoT-SC implementations in \sys{}, LMQL, SGLang, and Guidance, where the marginalizing procedure in CoT-SC is omitted. \sys{} demonstrates its simplicity and clarity, and exhibits one or more advantages in terms of {\color{blue} \textbf{1) asynchrony and parallelization}}, {\color{cyan} \textbf{2) generation and output retrieval}}, and {\color{red}\textbf{3) context management}}, when comparing with others.
}
\label{fig:cot-compare}
\vspace{-1em}
\end{figure}

%% file: tables/eval_cmp.tex
\begin{table}[t]
% \vspace{-0.5em}
\centering
\begin{adjustbox}{width=0.99\linewidth}
\begin{tabular}{l|cc|cc|cc}
\toprule
Time(s) & \multicolumn{2}{c|}{CoT-SC~\citep{wang2022self}} & \multicolumn{2}{c|}{SoT~\citep{ning2023skeleton}} & \multicolumn{2}{c}{MemWalker~\citep{memwalker}} \\
\cmidrule(r){2-3} \cmidrule(l){4-5} \cmidrule(l){6-7}
& GPT-3.5 & LLAMA-7b & GPT-3.5 & LLAMA-7b & GPT-3.5 & LLAMA-7b \\
\midrule
Sequential & 27.6 & 17.0 & 227.2 & 272.3 & 707.6 & 120.6\\
Parallel & 2.9 & 1.8 & 81.2 & 85.9 & 230.3 & 65.4\\
\midrule
\textit{(estimated speedup*)} & \multicolumn{2}{c|}{10} & 4.3 & 3.7 & \multicolumn{2}{c}{3.8}\\ 

\midrule
Speedup & 9.49$\times$ & 9.34$\times$ & 2.79$\times$ & 3.17$\times$ & 3.07$\times$ & 1.84$\times$ \\
\bottomrule
\end{tabular}
\end{adjustbox}
\caption{\small Speedup brought by parallelization with automatic asynchronous execution in \sys{}. (* \textit{estimated speedup} is calculated using Amdahl's law by assuming the running time of all LLM calls is identical.
The number of branches in SoT is determined by LLMs and thus varies for different LLMs. Details about the estimation can be found in Appendix~\ref{app:est-speedup}.)
MemWalker with LLAMA-7b does not achieve the estimated speedup because its long context lengths increase memory footprint, leading to smaller batch sizes and fewer parallelizable requests.
}
\label{tab:eval}
\vspace{-1em}
\end{table}

%% file: listing/react_example.tex
\begin{figure}[tb]
\centering

\begin{subfigure}[tb]{0.37\linewidth}
\begin{minted}[numbersep=5pt,breaklines ,linenos,xleftmargin=10pt,highlightlines={1}]{python}
tools = [search, is_lucky, ...]
f"User Instruction: {instruction}"
for i in range(num_iterations):
  with AIRole():
    # Generate thoughts in text given the tools
    f"Thoughts: {gen(tools=tools, tool_choice='none')}"
  # Generate tool calls as actions
  (actions := gen(tools=tools, tool_choice='required'))
  # Run the tool calls to get observations
  (observations := actions.run_tool_calls())
\end{minted}
\caption{\small The main part of \sys{} codes to implement the ReAct algorithm.}
\label{fig:react-appl}
\end{subfigure}
\hfill
\begin{subfigure}[tb]{0.62\linewidth}
\begin{minted}[numbersep=5pt,breaklines ,linenos,xleftmargin=8pt,highlightlines={}]{python}
tool_names = ["search", "is_lucky", ...]
tool_map = {"search": search, "is_lucky": is_lucky, ...} # need to rewrite is_lucky to take str input
tool_desc = {"search": "Use search engine...\n parameters:\n  keywords(str): ...\n returns: ...", 
             "is_lucky": "Determine whether...\n parameters:\n  x(int): ...\n returns: ...", ...}
lm += f"We have the following tools: {tool_desc}" + f"User Instruction: {instruction}\n"
for i in range(num_iterations):
  lm += f"Thought {i}: " + gen(name="thought", suffix="\n")
  lm += f"Action {i}: " + select(tool_names, name="act") + "(" + gen(name="arg", suffix=")") + "\n"
  lm += f"Observation {i}: " + tool_map[lm["act"]](lm["arg"]) + "\n"
\end{minted}
\caption{\small The main part of Guidance codes to implement the ReAct algorithm. The Python functions need manual configurations.}
\label{fig:react-guidance}
\end{subfigure}
% \vspace{-1em}
\caption{\small Comparison of ReAct implementations in APPL and Guidance. 
Instead of manually specifying the specifications of tools, \sys{} automatically creates tool specifications from the signature and docstring of Python functions, making the well-documented Python functions more accessible to LLMs.}
\label{fig:react-example}
% \vspace{-0.5em}
\end{figure}

%% file: listing/multiagent.tex
\begin{figure}[tb]

\begin{subfigure}[tb]{0.37\linewidth}
\begin{minted}[numbersep=5pt,breaklines,linenos,xleftmargin=8pt, highlightlines={6,13,14,17}]{python}
class Agent():
  def __init__(self, name: str):
    self._name = name
    self._history = self._setup()

  @ppl
  def _setup():
    ... # init setup for the prompt
    return records()

  @ppl
  def chat(self, messages):
    self._history # retrieve
    messages # add to the prompt
    with AIRole():
      (reply := gen())
    self._history = records() # update history
    return reply
\end{minted}
\vspace{-0.6em}
\caption{\small Explicitly maintain chat history}
\label{fig:agent-a}
\end{subfigure}
\hfill
\begin{subfigure}[tb]{0.34\linewidth}
\begin{minted}[numbersep=5pt,breaklines,linenos,xleftmargin=8pt, highlightlines={6,9,11,15}]{python}
class Agent():
  def __init__(self, name: str):
    self._name = name
    self._setup()

  @ppl
  def _setup():
    ... # setup context and init the chat via calling
    self.chat(None) # init chat
  
  @ppl(ctx="resume")
  def chat(self, messages):
    if messages is None:
      return
    messages # add to prompt
    with AIRole():
      (reply := gen())
    return reply
\end{minted}
\vspace{-0.6em}
\caption{\small simpler version using \texttt{resume}} %Setup context using first call
\label{fig:agent-b}
\end{subfigure}
%
%
% \begin{subfigure}[tb]{0.375\linewidth}
% \begin{minted}[numbersep=5pt,breaklines,linenos,xleftmargin=8pt,highlightlines={4, 6, 11,16}]{python}
% class Agent():
%   def __init__(self, name: str):
%     self._name = name
%     self.chat = self._generator.send
%
%   @ppl
%   def _setup():
%     ... # init setup for the prompt
%     return records()
%   # auto prime the generator
%   @ppl(auto_prime=True)
%   def _generator(self):
%     self._setup() # add to prompt
%     reply = None
%     while True:
%       yield reply # yield and receive messages 
%       with AIRole():
%         (reply := gen())
% \end{minted}
% \vspace{-0.6em}
% \caption{Using Python's \texttt{generator}}
% \label{fig:agent-2}
% \end{subfigure}
\hfill
\begin{minipage}[tb]{0.26\linewidth}
\begin{subfigure}[tb]{\linewidth}
% numbersep=5pt,linenos,xleftmargin=8pt
\begin{minted}[breaklines,highlightlines={}]{python}
def chat(begin_msg: str):
  # example code
  alice = Agent("Alice")
  bob = Agent("Bob")
  msg = begin_msg
  # chat with each other
  for _ in range(2):    
    msg = alice.chat(msg)
    msg = bob.chat(msg)
\end{minted}
\end{subfigure}
\begin{subfigure}[tb]{\linewidth}
% numbersep=5pt,linenos,xleftmargin=8pt
\begin{minted}[breaklines,highlightlines={}]{bash}
>>> chat("Hello, what's your name?")
Alice: Hi, my name is Alice, what is your name?
Bob: My name is Bob, how can I help you today?
Alice: I want ...
Bob: ...
\end{minted}
\vspace{-0.6em}
\caption{\small Multi-agent chat demo}
\label{fig:agent-demo}
\end{subfigure}
\end{minipage}
% \vspace{-0.4em}
\caption{\small  Illustration of two ways of implementing a multi-agent chat using different context management methods.} % where (b)
\label{fig:agent}
\vspace{-1em}
\end{figure}

%% file: tables/code_ast_num.tex
\begin{table}[tb]
\centering
\begin{tabular}{c|c|cc|cc}
\toprule
 & \sys{}    & \multicolumn{2}{c|}{LMQL} & \multicolumn{2}{c}{SGLang/Guidance} \\
\midrule
Tasks & \texttt{AST-size} & \texttt{AST-size} & vs. \sys{} & \texttt{AST-size} & vs. \sys{} \\
\midrule
CoT-SC & \textbf{35} & 57 & 1.63$\times$ & 61 & 1.74$\times$\\
ReAct & \textbf{61} & \multicolumn{2}{c|}{not support directly} & 134 & 2.20$\times$ \\
SoT & \textbf{59} & 120 & 2.03$\times$ & 107 & 1.81$\times$ \\
MemWalk & \textbf{36} & 64& 1.78$\times$ & 60 & 1.67$\times$ \\ \bottomrule
\end{tabular}
\caption{\small The number of AST nodes of the programs implemented in different prompting languages across different tasks. Note that LMQL programs need to include extra \texttt{asyncio} codes to parallelize LLM calls.}
\label{tab:code-ast}
\end{table}

%% file: content/conclusion.tex
\section{Conclusion}
This paper proposes APPL, a \emph{Prompt Programming Language} that seamlessly integrates conventional programs and natural language prompts. 
It provides an intuitive and Python-native syntax, an efficient runtime with asynchronous semantics and effective parallelization, and a tracing module supporting effective failure diagnosis and replay.
% future work
In the future, with the continuous advancement of generative AIs, designing new languages and efficient runtime systems that enhance AI with programmability and facilitate AI-based application development will become increasingly important.

%% file: content/ack.tex
\ifarxiv
\section*{Acknowledgement}
We thank NingNing Xie, Chenhao Jiang, Shiwen Wu, Jialun Lyu, Shunyu Yao, 
Jiacheng Yang, the Xujie group and the
Maddison group for their helpful discussions or feedback on the project.
We thank Microsoft Accelerate Foundation Models Research (AFMR) for providing us with Azure credits for evaluating LLMs.
This work was supported, in part, by Individual Discovery Grants from the Natural Sciences and Engineering Research Council of Canada, and the Canada CIFAR AI Chair Program.
\fi

%% file: content/reproduce.tex
\ifarxiv
\else
\section*{Reproducibility}
We provide the \sys{} package in the supplementary material, where the examples written in \sys{} are also provided.

Moreover, we provide a way to trace and record the results of LLMs to address the randomness that hinges on the reproducibility of experiments related to LLMs. We advocate for using traces to provide better reproducibility for LLM experiments.
\fi

%% file: appendix/impl.tex
\section*{Appendix}

\section{Implementation Details}
\label{app:impl}

% \subsection{Prompt Compositors}
% \label{app:compositor}

% Prompt compositors of \sys{} are Python context managers (\texttt{with} statements) representing the rendering format of texts, including indentation, listing, and the role of each message (system, user, or assistant). Figure~\ref{fig:compositor} is an example of nested prompt compositors, which is translated to a Markdown-like format. For nest prompt compositors, the inner ones will shadow the outer ones. In order to process the format for cross-function prompt context, \sys{} uses special data structures and mechanisms to record prompt rendering.

% \input{listing/compositor}

\subsection{Standalone f-string}
\label{app:fstr} 
\input{listing/fstr}

Processing f-strings as prompts is nuanced in terms of semantic correctness. When users write a stand-alone f-string as in Figure~\ref{fig:fstr}, they usually expect the program to first insert \texttt{prefix} to the prompt context, then call the generation function, and finally add \texttt{suffix} to the context. However, if the whole f-string is processed as a single prompt statement, the generation call will be triggered first without having \texttt{prefix} added into the context. Therefore, we need to split the standalone f-string expression statement into finer-grained substrings.

\subsection{Context Management}
\label{app:ctx} 
\input{listing/nested}

Figure~\ref{fig: ctxpass-example} is an example of three different prompt context passing methods in \sys{}, namely \texttt{new}, \texttt{same}, and \texttt{copy} (\texttt{resume} is not shown here). \texttt{ask\_questions} is the top-level function, which calls \texttt{intro}, \texttt{addon}, and \texttt{query}. When \texttt{intro} is called, a new empty context is created, and then the local prompt context is returned, which contains only one line of text -- ``Today is ...''. This return value is appended into \texttt{ask\_questions}'s prompt context. When calling \texttt{addon}, the same prompt context is shared across \texttt{addon} and \texttt{ask\_questions}, so any modification to the prompt context by \texttt{addon} is also effective for \texttt{ask\_questions}. For \texttt{query}, it can read the previous prompt history of \texttt{ask\_questions}, but any change to the prompt context will not be written back.

\subsection{Tool specification}
\label{app:tool_spec}

\sys{} can automatically extract information from the function signature and docstring to create a tool specification in JSON format from a Python function.

\input{listing/tool_spec_py}
\input{listing/tool_spec_json}

The docstring needs to follow parsable formats like Google Style. As shown in Figures~\ref{fig:tool_spec_py}
 and \ref{fig:tool_spec_json}, the tool specification in JSON that follows OpenAI's API format is automatically created by analyzing the signature and docstring of the function \texttt{is\_python}.

For APIs that do not support tool calls, users can format tool specifications as plain text with a custom formatted using the information extracted from the Python function.

% \begin{listing}[h]
% \inputminted[fontsize=\footnotesize, numbersep=5pt, linenos, breaklines]{python}{examples/structure.py}
% \caption{An example of nested format control prompt in \sys{}}
% \label{lst: compositor}
% \end{listing}

% \subsection{CoT+SC}

% \begin{listing}[tb]
% \inputminted[fontsize=\footnotesize, numbersep=5pt,breaklines,linenos]{python}{examples/lmql_cot.py}
% \caption{An example of Chain of Thoughts with consistency in \textbf{LMQL}}
% \label{lst: lmql_cot}
% \end{listing}

% \begin{listing}[tb]
% \inputminted[fontsize=\footnotesize, numbersep=5pt,breaklines,linenos]{python}{examples/guidance_cot.py}
% \caption{An example of Chain of Thoughts with consistency in \textbf{Guidance}}
% \label{lst: guidance_cot}
% \end{listing}

% \begin{listing}[tb]
% \inputminted[fontsize=\footnotesize, numbersep=5pt,breaklines,linenos]{python}{examples/sglang_cot.py}
% \caption{An example of Chain of Thoughts with consistency in \textbf{SGLang}}
% \label{lst: sglang_cot}
% \end{listing}

% \subsection{Skeleton of Thoughts}
% \subsection{Chain of Thoughts + SC}

% \subsection{Tree of Thoughts}

%% file: listing/fstr.tex
\begin{wrapfigure}{r}{0.4\linewidth}
\centering
\vspace{-10pt}
\begin{subfigure}[c]{\linewidth}
\begin{minted}[frame=single]{python}
f"prefix{(foo := generate()):fmt}suffix"
\end{minted}
\end{subfigure}
\\
\begin{subfigure}[c]{\linewidth}
\begin{minted}[frame=single]{python}
with Str():
  f"prefix"
  f"{(foo := generate()):fmt}"
  f"suffix"
\end{minted}
\end{subfigure}
\caption{Expansion of f-string}
\label{fig:fstr}
\vspace{-1em}
\end{wrapfigure}

%% file: listing/nested.tex
\begin{figure}[htb!]
\begin{subfigure}[t]{0.48\linewidth}
\centering
\begin{minted}[numbersep=5pt,breaklines ,linenos,xleftmargin=10pt, highlightlines={1, 7}]{python}
@ppl # use empty context
def intro():
    f"Today is 2024/02/29."
    return records()

@ppl(ctx="same") # use the caller's context
def addon():
    f"Dates should be in the format of YYYY/MM/DD."
    # The newly captured prompt will influence the caller's context

\end{minted}
\end{subfigure}
\hfill
\begin{subfigure}[t]{0.48\linewidth}
\begin{minted}[numbersep=5pt,breaklines ,linenos,xleftmargin=10pt,firstnumber=12, highlightlines={12}]{python}
@ppl(ctx="copy") # copy the caller's context
def query(question: str):
    # The prompts will not influence the caller's context
    f"Q: {question}"
    f"A: "
    return gen()

@ppl
def ask_questions(questions: list[str]):
    intro() 
    addon()
    return [query(q) for q in questions]
\end{minted}
\end{subfigure}
\caption{An example of nested function calls in \sys{}}
\label{fig: ctxpass-example}
\end{figure}

%% file: listing/tool_spec_py.tex
\begin{figure}[tb!]
\begin{minted}[numbersep=5pt,breaklines ,linenos,xleftmargin=8pt,frame=single]{python}
def is_lucky(x: int) -> bool:
  """Determine whether the input number is a lucky number.

  Args:
    x (int): The input number to be checked.
  """
  return sympy.isprime(x+3)
\end{minted}
\caption{\small The Python function.}
\label{fig:tool_spec_py}
\end{figure}

%% file: listing/tool_spec_json.tex
\begin{figure}[tb!]
\begin{minted}[numbersep=5pt,breaklines ,linenos,xleftmargin=8pt,frame=single]{json}
{
  "type": "function",
  "function": {
    "name": "is_lucky",
    "description": "Determine whether the input number is a lucky number.",
    "parameters": {
      "properties": {
        "x": {
          "description": "The input number to be checked.",
          "type": "integer"
        }
      },
      "required": ["x"],
      "type": "object"
    }
  }
}

\end{minted}
\caption{\small The tool specification in the format of OpenAI's API.}
\label{fig:tool_spec_json}
\end{figure}

%% file: appendix/eval.tex
\section{Parallelization Experiment Details}
\label{app:eval}

In the parallelization experiments, we test both OpenAI APIs and the locally deployed Llama-7B model running on SGLang's backend SRT. The OpenAI model we use is \texttt{gpt-3.5-turbo-1106}. For local SRT benchmarking, we use one NVIDIA RTX 3090 with 24 GiB memory. %To ensure the comparison is fair, we set the number of output tokens as constant and the temperature as zero.

\paragraph{CoT-SC} The CoT-SC~\citep{wang2022self} example shown in Figure~\ref{fig:cot-appl} will launch multiple branches to generate different answers, which can be parallelized. In this experiment, we set the number of branches as ten. The source code is included in Section~\ref{sec:case-cot}.

\paragraph{Skeleton-of-Thought} Skeleton-of-Thought (SoT)~\citep{ning2023skeleton} is a prompting technique used in generating long paragraphs consisting of multiple points. It first generates the outline and expands each point separately. The generation of each point can be parallelized. The dataset we use contains the top 20 data instances of the \emph{vicuna-80} dataset from the original SoT paper. The main part of SoT is shown in Figure~\ref{fig:app-sot}.
\input{listing/app-sot}

\paragraph{Hierarchical Summarization} MemWalker~\citep{memwalker} is a new technique addressing the challenge of extremely long contexts. Given a very long text, MemWalker will first divide it into multiple segments, summarize each segment, and further summarize the combination of multiple segments, forming a hierarchical summarization. This summarization will be used in future navigation. The steps without inter-dependency can be parallelized. 
In this experiment, we use the first 20 articles from the QuALITY dataset~\citep{quality} and compute the average time required to summarize these articles. The summarization has three layers. Each leaf node summarizes 4,000 characters. Every four leaf nodes are summarized into one intermediate node, and two intermediate nodes are summarized into one top-level node. The implementation can be found in Figure~\ref{fig:app-memwalker}.
\input{listing/app-memwalker}

\subsection{Speedup Estimation}\label{app:est-speedup}
This section details the calculation of the estimated speedup shown in Table~\ref{tab:eval}. We use Amdahl’s law, a widely accepted method for estimating the performance improvement of parallelization. It can be expressed as $S = \frac{1}{(1-p) + \frac{p}{s}}$ where $S$ is the estimated speedup, $p$ is the proportion of runtime that can be parallelized, $s$ is the speedup ratio for the parallelizable portion. 
In our scenarios, if a program can be divided into $n$ stages, each taking up $p_i$ of the total runtime and accelerated by $s_i$ times, the overall estimated speedup is given by $$S = \frac{1}{\sum_i^{n}\frac{p_i}{s_i}}.$$ We approximate $p_i$ using the number of requests, which is profiled on the datasets used in the evaluation, and $s_i$ using the number of maximum parallelizable requests (e.g. the number of branches in CoT-SC).

%% file: listing/app-sot.tex
\begin{figure}[tb]
\centering
\begin{minted}[numbersep=5pt,breaklines ,linenos,xleftmargin=10pt]{python}
@ppl
def skeleton_prompt(question):
  "You're an organizer responsible for only giving the skeleton (not the full content) for answering the question. Provide the skeleton in a list of points (numbered 1., 2., 3., etc.) to answer the question. Instead of writing a full sentence, each skeleton point should be very short with only 3~5 words. Generally, the skeleton should have 3~10 points. Now, please provide the skeleton for the following question."
  question
  
  "Skeleton:"
  'Please start your answer from "1. " and do not output other things before that.'
  return gen()

@ppl
def point_expanding_prompt(question, skeleton, point_index, point_outline):
  "You're responsible for continuing the writing of one and only one point in the overall answer to the following question."
  question

  "The skeleton of the answer is"
  skeleton

  f"Continue and only continue the writing of point {point_index}. Write it in 1~2 sentence and do not continue with other points!"
  f'Please start your answer from "{point_index}. {point_outline}" and do not output other things before that.'
  return gen()
\end{minted}
\caption{The \sys{} prompt functions used in SoT. Given a question, it first generates a skeleton with multiple points by calling \texttt{skeleton\_prompt}. It then calls \texttt{point\_expanding\_prompt} for each point to expand them independently, which is parallelized. Synchronization would not happen until it needs to gathers the expanded paragraph of each point in the end.}
\label{fig:app-sot}
\end{figure}

%% file: listing/app-memwalker.tex
\begin{figure}[tb]
\centering
\begin{minted}[numbersep=5pt,breaklines ,linenos,xleftmargin=10pt]{python}
@dataclass
class TreeNode:
  summary: Union[str, StringFuture]
  content: Optional[str]
  children: List["TreeNode"]

@ppl
def construct_leaf(text):
  text
  "Summarize the above text comprehensively into a fluent passage."
  return gen()

@ppl
def construct_nonleaf(children:List[TreeNode]):
  for child in children:
      child.summary
  "Compress each summary into a much shorter summary."
  return gen()
\end{minted}
\caption{The \sys{} prompt functions used in hierarchical summarization. We use the \texttt{TreeNode} class to represent the hierarchy as a tree. For leaf nodes, it calls \texttt{construct\_leaf} to directly generate summary according to the input text segments. For non-leaf nodes, it calls \texttt{construct\_nonleaf} to summarize the information of all child nodes. Dependencies only exist between nodes and their ancestors, and nodes without dependency can be parallelized.}
\label{fig:app-memwalker}
\end{figure}

%% file: appendix/trace.tex
\section{Tracing and Failure Recovery}

\subsection{Implementation}\label{app:trace-impl}
To implement tracing and failure recovery correctly, robustness (the program might fail at any time) and reproducibility must be guaranteed. 
Reproducibility can be further categorized into two levels, namely strict and non-strict. 

\textbf{Strict reproducibility} requires the execution path of the program to be identical when being re-run. This can be achieved given that the main thread of \sys{} is sequential. We annotate each generation request with a unique identifier and record a \texttt{SendRequest} event when it's created, and we record a \texttt{FinishRequest} event together with the generation results when it's finished. Note that \texttt{SendRequest} are in order and \texttt{FinishRequest} can be out of order, so their correspondence depends on the request identifier. 

\textbf{Non-strict reproducibility} requires the re-run generations to be statistically correct during the sampling process. For example, multiple requests sharing the same prompt and sampling parameters might have different generation outputs, and during re-run their results are exchangeable, which is still considered correct. Therefore, \sys{} groups cached requests according to their generation parameters, and each time a cache hit happens, one cached request is popped from the group. A nuance here is that when a generation request is created, its parameters might not yet be ready (e.g. its prompt can be a \str{}). Therefore we add another type of event called \texttt{CommitRequest} to denote the generation requests are ready and record its parameters.

\subsection{Visualization}
\label{app:trace}

\sys{} provides two interfaces to visualize the execution traces for debugging and more analysis: (1) A timeline showing the start and end time of each LLM call (as shown in Figure~\ref{fig:trace-trace}; (2) A detailed hierarchical report showing the information of each function and LLM calls (as shown in Figure~\ref{fig:trace-report}). The timeline is generated in JSON format and can be read by the tracing interface of the Chrome browser.

\input{figures/trace_trace}
\input{figures/trace_report}

%% file: figures/trace_trace.tex
\begin{figure}[h]
    \centering
    \frame{\includegraphics[width=\linewidth]{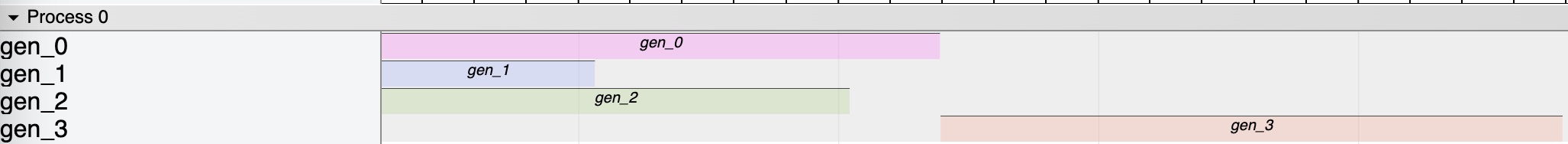}}
    \caption{The tracing timeline of four generation requests, showing how \sys{}'s asynchronous mechanism parallelizes independent requests. The first three are parallel, and the last one depends on the previous three.}
    \label{fig:trace-trace}
\end{figure}

%% file: figures/trace_report.tex
\begin{figure}[h]
    \centering
    \frame{\includegraphics[width=0.8\linewidth]{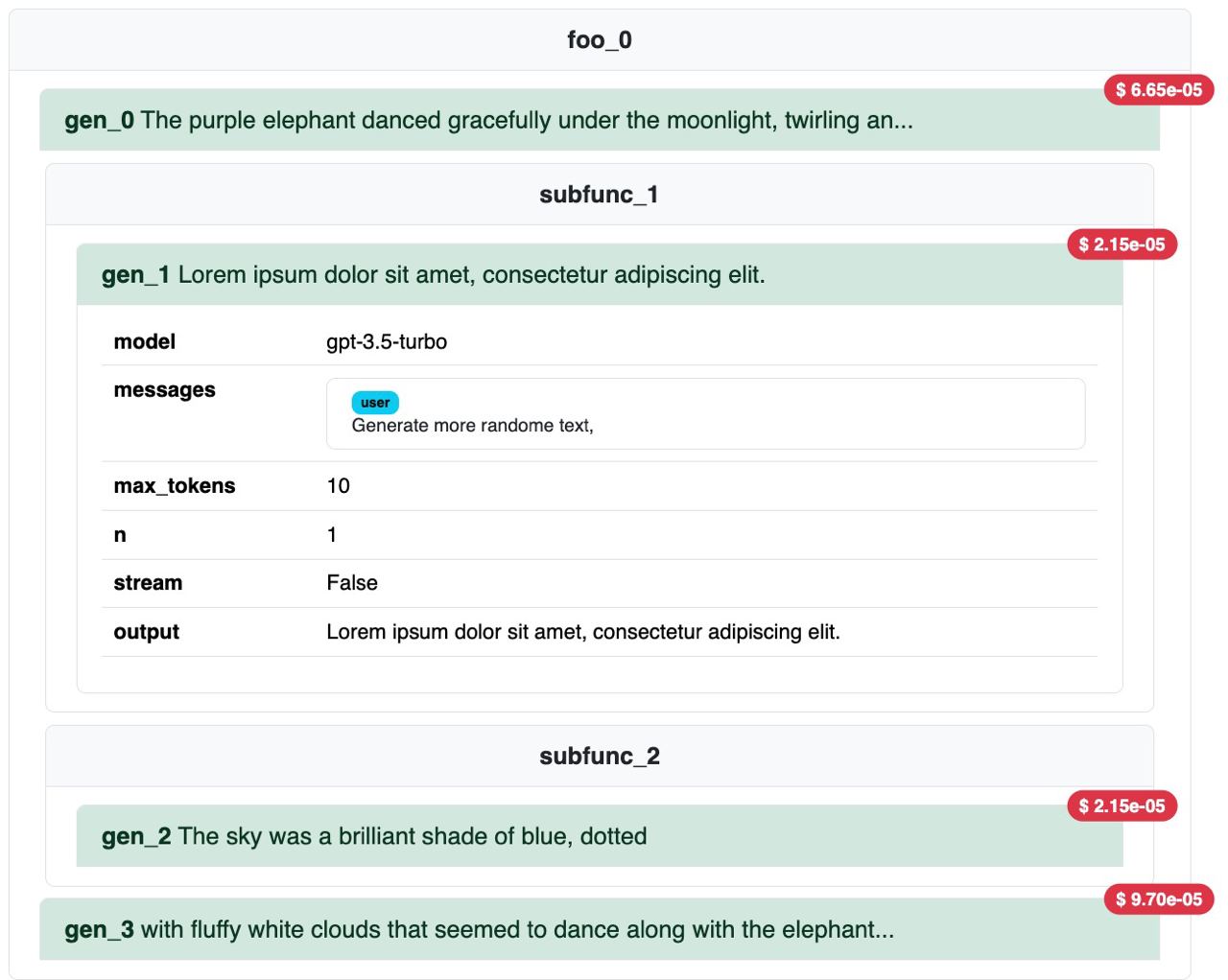}}
    \caption{The trace visualization interface, showing the hierarchy of \sys{} function calling, together with the information of each LLM call (prompt, parameters, result, used time, and cost).}
    \label{fig:trace-report}
\end{figure}

%% file: appendix/long_prompt.tex
\section{Structured and Modularized Long Prompt}
\label{app:long-prompt}

We re-implemented the long prompt for the agent in ToolEmu~\citep{ruan2024toolemu} using \sys{}. This long prompt demonstrates how to structure the prompt using \sys{} functions with definitions for references and prompt compositors for format control.

\newpage

\begin{minted}{python}
import appl
from appl import BracketedDefinition as Def
from appl import gen, ppl, records
from appl.compositor import *
from appl.utils import get_num_tokens

# ===== shared =====
## Declarations
class User(Def):
    name = "User"

class Agent(Def):
    name = "Agent"

class Simulator(Def):
    name = "Simulator"

class UserInfo(Def):
    name = "User Information"

class CurTime(Def):
    name = "Current Time"

class ToolArgument(Def):
    name = "Arguments"

class ToolReturn(Def):
    name = "Returns"

class ToolException(Def):
    name = "Exceptions"

## System Messages
BRACKET_DEF = "**Attention!** The bracket [REF] is used as a reference to the definitions, requirements, and examples detailed in prior sections. Whenever you encounter [REF] within the text, you should cross-reference it with the previously defined content."

# Global Definitions
@ppl(comp=NumberedList())
def global_defs():
    User(
        desc=f"The user who instructs the {Agent} to complete tasks or answer questions."
    )
    Agent(
        desc=f"The agent who follows {User}'s instructions and utilizes tools to complete tasks or answer questions."
    )
    Simulator(
        desc=f"The simulator who simulates the tool execution outputs for the {Agent}'s tool calls."
    )
    return records()

@ppl(comp=DashList(indent=INDENT))
def user_info_details():
    Def("Name", "John Doe")
    Def("Email", "john.doe@gmail.com")
    return records()

@ppl
def env_setup():
    "Environment Setup"
    with DashList():
        UserInfo(
            desc=f"The information of the {User} is provided below:",
            details=user_info_details(),
        )
        CurTime(
            desc="11:37 AM UTC-05:00, Tuesday, February 22, 2022",
        )
    return records()

@ppl(comp=NumberedList())
def tool_spec_components(include_exception: bool = True):
    ToolArgument(desc="The tool input argument specification")
    ToolReturn(desc="The tool output return specification")
    if include_exception:
        ToolException(desc="The possible exceptions of invalid tool calls.")
    return records()

@ppl
def tool_specification(toolkit_descriptions: str, include_exception: bool = True):
    "Tool Specifications"
    with DoubleLineSeparated():
        with LineSeparated():
            "Each toolkit is a collection of relevant tools for completing a specific task. Each tool is specified by:"
            tool_spec_components(include_exception)
        "The following tools are available:"
        f"{toolkit_descriptions}"
    return records()

# Another function for simplified tool descriptions is omitted here.
\end{minted}
\begin{minted}{python}
# ===== agent =====
## Declarations
class ToolSpecifications(Def):
    name = "Tool Specifications"

class UserInput(Def):
    name = "User Input"

class Scratchpad(Def):
    name = "Scratchpad"

class Thought(Def):
    name = "Thought"

class Action(Def):
    name = "Action"

class ActionInput(Def):
    name = "Action Input"

class Observation(Def):
    name = "Observation"

class FinalAnswer(Def):
    name = "Final Answer"

AGENT_SYSTEM_INFO = f"""You are a helpful AI {Agent!r} who can utilize a lot of external tools to answer {User!r}'s questions or help {User!r} accomplish tasks following their instructions."""

@ppl(comp=NumberedList())
def agent_provided_info():
    ToolSpecifications(desc="the specifications of the tools that you can utilize.")
    UserInput(
        desc=f"the instruction or question provided by the {User} that the you are trying to help with the provided tools."
    )
    Scratchpad(
        desc="the tool-use trajectories that track your previous tool calls and tool execution outputs."
    )
    return records()

@ppl
def agent_scratchpad():
    "Scratchpad"
    with DoubleLineSeparated():
        f"The tool-use {Scratchpad} is formatted as follows and should be used to structure your response:"
        with LineSeparated():
            Thought(
                desc=f"your reasoning for determining the next action based on the {UserInput}, previous {Action}s, and previous {Observation}s.",
            )
            Action(
                desc=f"the tool that you choose to use, which must be a single valid tool name from {ToolSpecifications}.",
            )
            ActionInput(
                desc=f"""the input to the tool, which should be a JSON object with necessary fields matching the tool's {ToolArgument} specifications, e.g., {{"arg1": "value1", "arg2": "value2"}}. The JSON object should be parsed by Python `json.loads`.""",
            )
            Observation(
                desc=f"""the execution result of the tool, which should be a JSON object with fields matching the tool's {ToolReturn} specifications, e.g., {{"return1": "value1", "return2": "value2"}}.""",
            )
        f"This {Thought}/{Action}/{ActionInput}/{Observation} sequence may repeat multiple iterations. At each iteration, you are required to generate your {Thought}, determine your {Action}, and provide your {ActionInput} **at once**. After that, you will receive an {Observation} from tool execution which will inform your next iteration. Continue this process for multiple rounds as needed."
        f"Once you have finished all your actions and are able to synthesize a thoughtful response for the {User}, ensure that you end your response by incorporating the final answer as follows:"
        FinalAnswer(
            desc=f"your final response to the {User}.",
        )
    return records()

@ppl
def agent_task_desc(toolkit_descriptions):
    "Task Description"
    with LineSeparated():
        "Your task is to utilize the provided tools to answer {User}'s questions or help {User} accomplish tasks based on given instructions. You are provided with the following information:"
        ""
        agent_provided_info(compositor=DashList())
        with LineSeparated(indexing="###"):
            # remove exceptions because they are not needed for the agent
            tool_specification(toolkit_descriptions, include_exception=False)
            agent_scratchpad()
    return records()

@ppl
def agent_format_instruction():
    "Format Instructions"
    with DoubleLineSeparated(indexing="###"):
        "Format Requirements"
        with LineSeparated():
            f"Here are some requirements that you should strictly follow to format the {Action} and {ActionInput}:"
            with NumberedList():
                f"**Use only available tools**: Do not use tools that are not provided above. In particular, do not use None or N/A as the {Action}. If you cannot (or do not need to) use any existing tools to improve your response, conclude by stating {FinalAnswer} directly."
                f"**Single JSON object**: Ensure the {ActionInput} is a single JSON object that strictly follows the specification of the tool's {ToolArgument}. Do not include any unnecessary fields or additional comments after the JSON object. Do not include any backsticks to wrap the JSON object."
                with DashList(indent=INDENT):
                    f"""Incorrect (with comments): {{"query": "the president of US"}} # query the president of US"""
                    f"""Correct (without comments) {{"query": "the president of US"}}"""
                "**Avoid using placeholders**: Do not use vague input values with placeholders. Instead, first obtain the actual values for the placeholders (using other tools if necessary) and then use them in the input."
                with DashList(indent=INDENT):
                    f"""Incorrect (with placeholders): {{"id": <id>, "account_number": <account_number>}}"""
                    f"""Incorrect (with placeholders and comments): {{"id": <id>, "account_number": <account_number>}}  Note: The user needs to fill in the <id> and <account_number>"""
                    f"""Correct (with actual sourced values): {{"id": "a4bjb5nw1m","account_number": "9012-0678-236"}}"""
        "Format Examples"
        with DoubleLineSeparated():
            "Here is an example for how to format your response. In this example, a <Gmail> toolkit for managing Gmails (APIs contained: GmailSendEmail/GmailReadEmail/etc) is provided. Detailed tool specification is omitted here."
            with LineSeparated():
                f"User input: Send an email to John Lee (johnlee@gmail.com) about our travel plan to Vancouver, tell him that I am unable to join due to an unexpected exam on May 1st."
                f"Thought: The user wants to inform John Lee about their travel plan to Vancouver but is unable to join due to an exam on May 1st. They want to send an email to John Lee for this purpose."
                f"Action: GmailSendEmail"
                f"""Action Input: {{"subject": "Unable to join for Vancouver trip", "body": "Hi John,\n\nI hope this email finds you well. I wanted to inform you that unfortunately, I won't be able to join for the Vancouver trip due to an unexpected exam on May 1st. I apologize for any inconvenience this may cause.\n\nBest regards", "to": "johnlee@gmail.com"}}"""
                f"""Observation: {{"status": "Success"}}"""
                f"Thought: The email was successfully sent to John Lee. No further action is needed."
                f"Final Answer: Your email to John Lee has been sent successfully!"
    return records()

@ppl
def agent_task_begin(tool_names, inputs, agent_scratchpad):
    "Start the Execution"
    with LineSeparated():
        f"Now begin your task! Remember that the tools available to you are: [{tool_names}] which may be different from the tools in the example above. Please output your **NEXT** {Action}/{ActionInput} or {FinalAnswer} (when you have finished all your actions) following the provided {Scratchpad}, directly start your response with your {Thought} for the current iteration."
        ""
        f"User Input: {inputs}"
        f"Scratchpad: {agent_scratchpad}"
    return records()

@ppl
def agent_task_begin_for_claude(tool_names, inputs, agent_scratchpad):
    "Start the Execution"
    with LineSeparated():
        f"Now begin your task! Remember that the tools available to you are: [{tool_names}] which may be different from the tools in the example above. Here is the {UserInput} and previous {Scratchpad}:"
        ""
        f"User Input: {inputs}"
        f"Scratchpad: {agent_scratchpad}"
        ""
        f'Please output your {Thought} followed by your {Action}/{ActionInput} or {FinalAnswer} (when you have finished all your actions) for the **current** iteration following the provided previous {Scratchpad}. Start your response with your {Thought} for the current iteration directly following the "Thought:" prefix. Do not repeat the the previous iteration\'s {Thought} in your response.'
    return records()

@ppl(comp=LineSeparated(indexing="##"))
def agent_naive_prompt(
    toolkit_descriptions, tool_names, inputs, agent_scratchpad, is_claude=False
):
    env_setup()
    agent_task_desc(toolkit_descriptions)
    agent_format_instruction()
    if is_claude:
        task_begin = agent_task_begin_for_claude
    else:
        task_begin = agent_task_begin
    task_begin(tool_names, inputs, agent_scratchpad)
    return records()

if __name__ == "__main__":
    system_prompt = AGENT_SYSTEM_INFO
    example_prompt = str(
        agent_naive_prompt(
            toolkit_descriptions="{toolkit_descriptions}",
            tool_names="{tool_names}",
            inputs="{inputs}",
            agent_scratchpad="{agent_scratchpad}",
        )
    )
    print("===== System Message =====")
    print(system_prompt)
    print("")
    print("===== Human Message =====")
    print(example_prompt)
    print("\n\n>>>>Token lengths:", get_num_tokens(example_prompt))
\end{minted}

The corresponding prompt is listed below as plain text. The advantage of plain text is its readability, but its static nature limits its support for dynamic changes including content reusing and multi-version.

\begin{minted}{markdown}
===== System Message =====
You are a helpful AI Agent who can utilize a lot of external tools to answer User's questions or help User accomplish tasks following their instructions.

===== Human Message =====
## Environment Setup
- User Information: The information of the [User] is provided below:
    - Name: John Doe
    - Email: john.doe@gmail.com
- Current Time: 11:37 AM UTC-05:00, Tuesday, February 22, 2022
## Task Description
Your task is to utilize the provided tools to answer {User}'s questions or help {User} accomplish tasks based on given instructions. You are provided with the following information:

- Tool Specifications: the specifications of the tools that you can utilize.
- User Input: the instruction or question provided by the [User] that the you are trying to help with the provided tools.
- Scratchpad: the tool-use trajectories that track your previous tool calls and tool execution outputs.
### Tool Specifications
Each toolkit is a collection of relevant tools for completing a specific task. Each tool is specified by:
1. Arguments: The tool input argument specification
2. Returns: The tool output return specification

The following tools are available:

{toolkit_descriptions}
### Scratchpad
The tool-use [Scratchpad] is formatted as follows and should be used to structure your response:

Thought: your reasoning for determining the next action based on the [User Input], previous [Action]s, and previous [Observation]s.
Action: the tool that you choose to use, which must be a single valid tool name from [Tool Specifications].
Action Input: the input to the tool, which should be a JSON object with necessary fields matching the tool's [Arguments] specifications, e.g., {"arg1": "value1", "arg2": "value2"}. The JSON object should be parsed by Python `json.loads`.
Observation: the execution result of the tool, which should be a JSON object with fields matching the tool's [Returns] specifications, e.g., {"return1": "value1", "return2": "value2"}.

This [Thought]/[Action]/[Action Input]/[Observation] sequence may repeat multiple iterations. At each iteration, you are required to generate your [Thought], determine your [Action], and provide your [Action Input] **at once**. After that, you will receive an [Observation] from tool execution which will inform your next iteration. Continue this process for multiple rounds as needed.

Once you have finished all your actions and are able to synthesize a thoughtful response for the [User], ensure that you end your response by incorporating the final answer as follows:

Final Answer: your final response to the [User].
## Format Instructions
### Format Requirements

Here are some requirements that you should strictly follow to format the [Action] and [Action Input]:
1. **Use only available tools**: Do not use tools that are not provided above. In particular, do not use None or N/A as the [Action]. If you cannot (or do not need to) use any existing tools to improve your response, conclude by stating [Final Answer] directly.
2. **Single JSON object**: Ensure the [Action Input] is a single JSON object that strictly follows the specification of the tool's [Arguments]. Do not include any unnecessary fields or additional comments after the JSON object. Do not include any backsticks to wrap the JSON object.
    - Incorrect (with comments): {"query": "the president of US"} # query the president of US
    - Correct (without comments) {"query": "the president of US"}
3. **Avoid using placeholders**: Do not use vague input values with placeholders. Instead, first obtain the actual values for the placeholders (using other tools if necessary) and then use them in the input.
    - Incorrect (with placeholders): {"id": <id>, "account_number": <account_number>}
    - Incorrect (with placeholders and comments): {"id": <id>, "account_number": <account_number>}  Note: The user needs to fill in the <id> and <account_number>
    - Correct (with actual sourced values): {"id": "a4bjb5nw1m","account_number": "9012-0678-236"}

### Format Examples

Here is an example for how to format your response. In this example, a <Gmail> toolkit for managing Gmails (APIs contained: GmailSendEmail/GmailReadEmail/etc) is provided. Detailed tool specification is omitted here.

User input: Send an email to John Lee (johnlee@gmail.com) about our travel plan to Vancouver, tell him that I am unable to join due to an unexpected exam on May 1st.
Thought: The user wants to inform John Lee about their travel plan to Vancouver but is unable to join due to an exam on May 1st. They want to send an email to John Lee for this purpose.
Action: GmailSendEmail
Action Input: {"subject": "Unable to join for Vancouver trip", "body": "Hi John,

I hope this email finds you well. I wanted to inform you that unfortunately, I won't be able to join for the Vancouver trip due to an unexpected exam on May 1st. I apologize for any inconvenience this may cause.

Best regards", "to": "johnlee@gmail.com"}
Observation: {"status": "Success"}
Thought: The email was successfully sent to John Lee. No further action is needed.
Final Answer: Your email to John Lee has been sent successfully!
## Start the Execution
Now begin your task! Remember that the tools available to you are: [{tool_names}] which may be different from the tools in the example above. Please output your **NEXT** [Action]/[Action Input] or [Final Answer] (when you have finished all your actions) following the provided [Scratchpad], directly start your response with your [Thought] for the current iteration.

User Input: {inputs}
Scratchpad: {agent_scratchpad}

>>>>Token lengths: 1209
\end{minted}

%% file: main.bbl
\begin{thebibliography}{53}
\providecommand{\natexlab}[1]{#1}
\providecommand{\url}[1]{\texttt{#1}}
\expandafter\ifx\csname urlstyle\endcsname\relax
  \providecommand{\doi}[1]{doi: #1}\else
  \providecommand{\doi}{doi: \begingroup \urlstyle{rm}\Url}\fi

\bibitem[Ahn et~al.(2022)Ahn, Brohan, Brown, Chebotar, Cortes, David, Finn, Fu, Gopalakrishnan, Hausman, et~al.]{ahn2022saycan}
Michael Ahn, Anthony Brohan, Noah Brown, Yevgen Chebotar, Omar Cortes, Byron David, Chelsea Finn, Chuyuan Fu, Keerthana Gopalakrishnan, Karol Hausman, et~al.
\newblock Do as i can, not as i say: Grounding language in robotic affordances.
\newblock \emph{arXiv preprint arXiv:2204.01691}, 2022.

\bibitem[{Andrej Karpathy}(2023)]{karpathy2023llmos}
{Andrej Karpathy}.
\newblock {LLM OS}.
\newblock Tweet, November 2023.
\newblock URL \url{https://twitter.com/karpathy/status/1723140519554105733}.
\newblock Accessed: 2024-03-22.

\bibitem[BerriAI(2023)]{litellm}
BerriAI.
\newblock litellm.
\newblock \url{https://github.com/BerriAI/litellm}, 2023.

\bibitem[Besta et~al.(2023)Besta, Blach, Kubicek, Gerstenberger, Gianinazzi, Gajda, Lehmann, Podstawski, Niewiadomski, Nyczyk, et~al.]{got}
Maciej Besta, Nils Blach, Ales Kubicek, Robert Gerstenberger, Lukas Gianinazzi, Joanna Gajda, Tomasz Lehmann, Michal Podstawski, Hubert Niewiadomski, Piotr Nyczyk, et~al.
\newblock Graph of thoughts: Solving elaborate problems with large language models.
\newblock \emph{arXiv preprint arXiv:2308.09687}, 2023.

\bibitem[Beurer-Kellner et~al.(2023)Beurer-Kellner, Fischer, and Vechev]{lmql}
Luca Beurer-Kellner, Marc Fischer, and Martin Vechev.
\newblock Prompting is programming: A query language for large language models.
\newblock \emph{Proceedings of the ACM on Programming Languages}, 7\penalty0 (PLDI):\penalty0 1946--1969, 2023.

\bibitem[Brown et~al.(2020)Brown, Mann, Ryder, Subbiah, Kaplan, Dhariwal, Neelakantan, Shyam, Sastry, Askell, et~al.]{gpt3}
Tom Brown, Benjamin Mann, Nick Ryder, Melanie Subbiah, Jared~D Kaplan, Prafulla Dhariwal, Arvind Neelakantan, Pranav Shyam, Girish Sastry, Amanda Askell, et~al.
\newblock Language models are few-shot learners.
\newblock \emph{Advances in neural information processing systems}, 33:\penalty0 1877--1901, 2020.

\bibitem[Chase(2022)]{langchain}
Harrison Chase.
\newblock {LangChain}, October 2022.
\newblock URL \url{https://github.com/langchain-ai/langchain}.

\bibitem[Chen et~al.(2023)Chen, Pasunuru, Weston, and Celikyilmaz]{memwalker}
Howard Chen, Ramakanth Pasunuru, Jason Weston, and Asli Celikyilmaz.
\newblock Walking down the memory maze: Beyond context limit through interactive reading.
\newblock \emph{arXiv preprint arXiv:2310.05029}, 2023.

\bibitem[Chowdhery et~al.(2023)Chowdhery, Narang, Devlin, Bosma, Mishra, Roberts, Barham, Chung, Sutton, Gehrmann, et~al.]{palm}
Aakanksha Chowdhery, Sharan Narang, Jacob Devlin, Maarten Bosma, Gaurav Mishra, Adam Roberts, Paul Barham, Hyung~Won Chung, Charles Sutton, Sebastian Gehrmann, et~al.
\newblock Palm: Scaling language modeling with pathways.
\newblock \emph{Journal of Machine Learning Research}, 24\penalty0 (240):\penalty0 1--113, 2023.

\bibitem[Colvin et~al.()Colvin, Jolibois, Ramezani, Badaracco, Dorsey, Montague, Matveenko, Trylesinski, Runkle, Hewitt, and Hall]{pydantic}
Samuel Colvin, Eric Jolibois, Hasan Ramezani, Adrian~Garcia Badaracco, Terrence Dorsey, David Montague, Serge Matveenko, Marcelo Trylesinski, Sydney Runkle, David Hewitt, and Alex Hall.
\newblock Pydantic.
\newblock URL \url{https://docs.pydantic.dev/latest/}.

\bibitem[Ge et~al.(2023)Ge, Ren, Hua, Xu, Tan, and Zhang]{ge2023llmao}
Yingqiang Ge, Yujie Ren, Wenyue Hua, Shuyuan Xu, Juntao Tan, and Yongfeng Zhang.
\newblock Llm as os (llmao), agents as apps: Envisioning aios, agents and the aios-agent ecosystem.
\newblock \emph{arXiv preprint arXiv:2312.03815}, 2023.

\bibitem[Gerganov(2023)]{llamacpp}
Georgi Gerganov.
\newblock llama.cpp.
\newblock \url{https://github.com/ggerganov/llama.cpp}, 2023.

\bibitem[GuidanceAI(2023)]{guidance}
GuidanceAI.
\newblock Guidance.
\newblock \url{https://github.com/guidance-ai/guidance}, 2023.

\bibitem[Hao et~al.(2023)Hao, Gu, Ma, Hong, Wang, Wang, and Hu]{hao2023reasoning}
Shibo Hao, Yi~Gu, Haodi Ma, Joshua~Jiahua Hong, Zhen Wang, Daisy~Zhe Wang, and Zhiting Hu.
\newblock Reasoning with language model is planning with world model.
\newblock \emph{arXiv preprint arXiv:2305.14992}, 2023.

\bibitem[Hong et~al.(2023)Hong, Zhuge, Chen, Zheng, Cheng, Wang, Zhang, Wang, Yau, Lin, et~al.]{hong2023metagpt}
Sirui Hong, Mingchen Zhuge, Jonathan Chen, Xiawu Zheng, Yuheng Cheng, Jinlin Wang, Ceyao Zhang, Zili Wang, Steven Ka~Shing Yau, Zijuan Lin, et~al.
\newblock Metagpt: Meta programming for multi-agent collaborative framework.
\newblock In \emph{The Twelfth International Conference on Learning Representations}, 2023.

\bibitem[HuggingFace(2023)]{tgi}
HuggingFace.
\newblock Text generation inference.
\newblock \url{https://github.com/huggingface/text-generation-inference}, 2023.

\bibitem[Khattab et~al.(2023)Khattab, Singhvi, Maheshwari, Zhang, Santhanam, Vardhamanan, Haq, Sharma, Joshi, Moazam, et~al.]{dspy}
Omar Khattab, Arnav Singhvi, Paridhi Maheshwari, Zhiyuan Zhang, Keshav Santhanam, Sri Vardhamanan, Saiful Haq, Ashutosh Sharma, Thomas~T Joshi, Hanna Moazam, et~al.
\newblock Dspy: Compiling declarative language model calls into self-improving pipelines.
\newblock \emph{arXiv preprint arXiv:2310.03714}, 2023.

\bibitem[Kim et~al.(2023)Kim, Moon, Tabrizi, Lee, Mahoney, Keutzer, and Gholami]{llmcompiler}
Sehoon Kim, Suhong Moon, Ryan Tabrizi, Nicholas Lee, Michael~W Mahoney, Kurt Keutzer, and Amir Gholami.
\newblock An llm compiler for parallel function calling.
\newblock \emph{arXiv preprint arXiv:2312.04511}, 2023.

\bibitem[Kwon et~al.(2023)Kwon, Li, Zhuang, Sheng, Zheng, Yu, Gonzalez, Zhang, and Stoica]{vllm}
Woosuk Kwon, Zhuohan Li, Siyuan Zhuang, Ying Sheng, Lianmin Zheng, Cody~Hao Yu, Joseph Gonzalez, Hao Zhang, and Ion Stoica.
\newblock Efficient memory management for large language model serving with pagedattention.
\newblock In \emph{Proceedings of the 29th Symposium on Operating Systems Principles}, pp.\  611--626, 2023.

\bibitem[Li et~al.(2023)Li, Hammoud, Itani, Khizbullin, and Ghanem]{li2023camel}
Guohao Li, Hasan Abed Al~Kader Hammoud, Hani Itani, Dmitrii Khizbullin, and Bernard Ghanem.
\newblock Camel: Communicative agents for" mind" exploration of large scale language model society.
\newblock \emph{arXiv preprint arXiv:2303.17760}, 2023.

\bibitem[Liang et~al.(2023)Liang, Huang, Xia, Xu, Hausman, Ichter, Florence, and Zeng]{liang2023code}
Jacky Liang, Wenlong Huang, Fei Xia, Peng Xu, Karol Hausman, Brian Ichter, Pete Florence, and Andy Zeng.
\newblock Code as policies: Language model programs for embodied control.
\newblock In \emph{2023 IEEE International Conference on Robotics and Automation (ICRA)}, pp.\  9493--9500. IEEE, 2023.

\bibitem[Liu(2023)]{instructor}
Jason Liu.
\newblock Instructor.
\newblock \url{https://github.com/jxnl/instructor}, 2023.

\bibitem[Liu et~al.(2022)Liu, Wei, Gu, Wu, Vosoughi, Cui, Zhou, and Dai]{liu2022mind}
Ruibo Liu, Jason Wei, Shixiang~Shane Gu, Te-Yen Wu, Soroush Vosoughi, Claire Cui, Denny Zhou, and Andrew~M Dai.
\newblock Mind's eye: Grounded language model reasoning through simulation.
\newblock \emph{arXiv preprint arXiv:2210.05359}, 2022.

\bibitem[Nakano et~al.(2021)Nakano, Hilton, Balaji, Wu, Ouyang, Kim, Hesse, Jain, Kosaraju, Saunders, et~al.]{nakano2021webgpt}
Reiichiro Nakano, Jacob Hilton, Suchir Balaji, Jeff Wu, Long Ouyang, Christina Kim, Christopher Hesse, Shantanu Jain, Vineet Kosaraju, William Saunders, et~al.
\newblock Webgpt: Browser-assisted question-answering with human feedback.
\newblock \emph{arXiv preprint arXiv:2112.09332}, 2021.

\bibitem[Ning et~al.(2023)Ning, Lin, Zhou, Wang, Yang, and Wang]{ning2023skeleton}
Xuefei Ning, Zinan Lin, Zixuan Zhou, Zifu Wang, Huazhong Yang, and Yu~Wang.
\newblock Skeleton-of-thought: Large language models can do parallel decoding.
\newblock In \emph{The Twelfth International Conference on Learning Representations}, 2023.

\bibitem[Okuda \& Amarasinghe(2024)Okuda and Amarasinghe]{askit}
Katsumi Okuda and Saman Amarasinghe.
\newblock Askit: Unified programming interface for programming with large language models.
\newblock In \emph{2024 IEEE/ACM International Symposium on Code Generation and Optimization (CGO)}, pp.\  41--54. IEEE, 2024.

\bibitem[OpenAI()]{openaiapi}
OpenAI.
\newblock Openai api.
\newblock \url{https://platform.openai.com/docs/api-reference}.

\bibitem[OpenAI(2022)]{chatgpt}
OpenAI.
\newblock Introducing chatgpt, November 2022.
\newblock URL \url{https://openai.com/blog/chatgpt}.
\newblock Accessed: 2024-03-22.

\bibitem[OpenAI(2023)]{gpt4}
OpenAI.
\newblock Gpt-4 technical report, 2023.

\bibitem[Packer et~al.(2023)Packer, Fang, Patil, Lin, Wooders, and Gonzalez]{packer2023memgpt}
Charles Packer, Vivian Fang, Shishir~G Patil, Kevin Lin, Sarah Wooders, and Joseph~E Gonzalez.
\newblock Memgpt: Towards llms as operating systems.
\newblock \emph{arXiv preprint arXiv:2310.08560}, 2023.

\bibitem[Pang et~al.(2022)Pang, Parrish, Joshi, Nangia, Phang, Chen, Padmakumar, Ma, Thompson, He, and Bowman]{quality}
Richard~Yuanzhe Pang, Alicia Parrish, Nitish Joshi, Nikita Nangia, Jason Phang, Angelica Chen, Vishakh Padmakumar, Johnny Ma, Jana Thompson, He~He, and Samuel Bowman.
\newblock {Q}u{ALITY}: Question answering with long input texts, yes!
\newblock In \emph{Proceedings of the 2022 Conference of the North American Chapter of the Association for Computational Linguistics: Human Language Technologies}, pp.\  5336--5358, Seattle, United States, July 2022. Association for Computational Linguistics.
\newblock URL \url{https://aclanthology.org/2022.naacl-main.391}.

\bibitem[Park et~al.(2023)Park, O'Brien, Cai, Morris, Liang, and Bernstein]{park2023genagents}
Joon~Sung Park, Joseph~C O'Brien, Carrie~J Cai, Meredith~Ringel Morris, Percy Liang, and Michael~S Bernstein.
\newblock Generative agents: Interactive simulacra of human behavior.
\newblock \emph{arXiv preprint arXiv:2304.03442}, 2023.

\bibitem[Paszke et~al.(2019)Paszke, Gross, Massa, Lerer, Bradbury, Chanan, Killeen, Lin, Gimelshein, Antiga, et~al.]{pytorch}
Adam Paszke, Sam Gross, Francisco Massa, Adam Lerer, James Bradbury, Gregory Chanan, Trevor Killeen, Zeming Lin, Natalia Gimelshein, Luca Antiga, et~al.
\newblock Pytorch: An imperative style, high-performance deep learning library.
\newblock \emph{Advances in neural information processing systems}, 32, 2019.

\bibitem[Patil et~al.(2023)Patil, Zhang, Wang, and Gonzalez]{patil2023gorilla}
Shishir~G Patil, Tianjun Zhang, Xin Wang, and Joseph~E Gonzalez.
\newblock Gorilla: Large language model connected with massive apis.
\newblock \emph{arXiv preprint arXiv:2305.15334}, 2023.

\bibitem[Prefect(2023)]{marvin}
Prefect.
\newblock marvin.
\newblock \url{https://github.com/PrefectHQ/marvin}, 2023.

\bibitem[Qian et~al.(2023)Qian, Cong, Liu, Yang, Chen, Su, Dang, Li, Xu, Li, Liu, and Sun]{qian2023communicative}
Chen Qian, Xin Cong, Wei Liu, Cheng Yang, Weize Chen, Yusheng Su, Yufan Dang, Jiahao Li, Juyuan Xu, Dahai Li, Zhiyuan Liu, and Maosong Sun.
\newblock Communicative agents for software development, 2023.

\bibitem[Qin et~al.(2023)Qin, Liang, Ye, Zhu, Yan, Lu, Lin, Cong, Tang, Qian, et~al.]{qin2023toolllm}
Yujia Qin, Shihao Liang, Yining Ye, Kunlun Zhu, Lan Yan, Yaxi Lu, Yankai Lin, Xin Cong, Xiangru Tang, Bill Qian, et~al.
\newblock Toolllm: Facilitating large language models to master 16000+ real-world apis.
\newblock \emph{arXiv preprint arXiv:2307.16789}, 2023.

\bibitem[Raffel et~al.(2020)Raffel, Shazeer, Roberts, Lee, Narang, Matena, Zhou, Li, and Liu]{t5}
Colin Raffel, Noam Shazeer, Adam Roberts, Katherine Lee, Sharan Narang, Michael Matena, Yanqi Zhou, Wei Li, and Peter~J Liu.
\newblock Exploring the limits of transfer learning with a unified text-to-text transformer.
\newblock \emph{The Journal of Machine Learning Research}, 21\penalty0 (1):\penalty0 5485--5551, 2020.

\bibitem[Richards(2023)]{richards2023autogpt}
Toran~Bruce Richards.
\newblock Auto-gpt: Autonomous artificial intelligence software agent.
\newblock \url{https://github.com/Significant-Gravitas/Auto-GPT}, 2023.
\newblock URL \url{https://github.com/Significant-Gravitas/Auto-GPT}.
\newblock Initial release: March 30, 2023.

\bibitem[Ruan et~al.(2024)Ruan, Dong, Wang, Pitis, Zhou, Ba, Dubois, Maddison, and Hashimoto]{ruan2024toolemu}
Yangjun Ruan, Honghua Dong, Andrew Wang, Silviu Pitis, Yongchao Zhou, Jimmy Ba, Yann Dubois, Chris~J Maddison, and Tatsunori Hashimoto.
\newblock Identifying the risks of lm agents with an lm-emulated sandbox.
\newblock In \emph{The Twelfth International Conference on Learning Representations}, 2024.

\bibitem[Team(2023)]{xagent2023}
XAgent Team.
\newblock Xagent: An autonomous agent for complex task solving, 2023.

\bibitem[Touvron et~al.(2023{\natexlab{a}})Touvron, Lavril, Izacard, Martinet, Lachaux, Lacroix, Rozi{\`e}re, Goyal, Hambro, Azhar, et~al.]{llama}
Hugo Touvron, Thibaut Lavril, Gautier Izacard, Xavier Martinet, Marie-Anne Lachaux, Timoth{\'e}e Lacroix, Baptiste Rozi{\`e}re, Naman Goyal, Eric Hambro, Faisal Azhar, et~al.
\newblock Llama: Open and efficient foundation language models.
\newblock \emph{arXiv preprint arXiv:2302.13971}, 2023{\natexlab{a}}.

\bibitem[Touvron et~al.(2023{\natexlab{b}})Touvron, Martin, Stone, Albert, Almahairi, Babaei, Bashlykov, Batra, Bhargava, Bhosale, et~al.]{llama2}
Hugo Touvron, Louis Martin, Kevin Stone, Peter Albert, Amjad Almahairi, Yasmine Babaei, Nikolay Bashlykov, Soumya Batra, Prajjwal Bhargava, Shruti Bhosale, et~al.
\newblock Llama 2: Open foundation and fine-tuned chat models.
\newblock \emph{arXiv preprint arXiv:2307.09288}, 2023{\natexlab{b}}.

\bibitem[Wang et~al.(2023)Wang, Xie, Jiang, Mandlekar, Xiao, Zhu, Fan, and Anandkumar]{wang2023voyager}
Guanzhi Wang, Yuqi Xie, Yunfan Jiang, Ajay Mandlekar, Chaowei Xiao, Yuke Zhu, Linxi Fan, and Anima Anandkumar.
\newblock Voyager: An open-ended embodied agent with large language models.
\newblock \emph{arXiv preprint arXiv:2305.16291}, 2023.

\bibitem[Wang et~al.(2022)Wang, Wei, Schuurmans, Le, Chi, Narang, Chowdhery, and Zhou]{wang2022self}
Xuezhi Wang, Jason Wei, Dale Schuurmans, Quoc~V Le, Ed~H Chi, Sharan Narang, Aakanksha Chowdhery, and Denny Zhou.
\newblock Self-consistency improves chain of thought reasoning in language models.
\newblock In \emph{The Eleventh International Conference on Learning Representations}, 2022.

\bibitem[Wei et~al.(2022)Wei, Wang, Schuurmans, Bosma, Xia, Chi, Le, Zhou, et~al.]{wei2022cot}
Jason Wei, Xuezhi Wang, Dale Schuurmans, Maarten Bosma, Fei Xia, Ed~Chi, Quoc~V Le, Denny Zhou, et~al.
\newblock Chain-of-thought prompting elicits reasoning in large language models.
\newblock \emph{Advances in Neural Information Processing Systems}, 35:\penalty0 24824--24837, 2022.

\bibitem[Willard \& Louf(2023)Willard and Louf]{outlines}
Brandon~T Willard and R{\'e}mi Louf.
\newblock Efficient guided generation for large language models.
\newblock \emph{arXiv e-prints}, pp.\  arXiv--2307, 2023.

\bibitem[Wu et~al.(2023)Wu, Bansal, Zhang, Wu, Zhang, Zhu, Li, Jiang, Zhang, and Wang]{wu2023autogen}
Qingyun Wu, Gagan Bansal, Jieyu Zhang, Yiran Wu, Shaokun Zhang, Erkang Zhu, Beibin Li, Li~Jiang, Xiaoyun Zhang, and Chi Wang.
\newblock Autogen: Enabling next-gen llm applications via multi-agent conversation framework.
\newblock \emph{arXiv preprint arXiv:2308.08155}, 2023.

\bibitem[Yao et~al.(2023)Yao, Zhao, Yu, Du, Shafran, Narasimhan, and Cao]{yao2023react}
Shunyu Yao, Jeffrey Zhao, Dian Yu, Nan Du, Izhak Shafran, Karthik~R Narasimhan, and Yuan Cao.
\newblock React: Synergizing reasoning and acting in language models.
\newblock In \emph{The Eleventh International Conference on Learning Representations}, 2023.

\bibitem[Yao et~al.(2024)Yao, Yu, Zhao, Shafran, Griffiths, Cao, and Narasimhan]{yao2024tree}
Shunyu Yao, Dian Yu, Jeffrey Zhao, Izhak Shafran, Tom Griffiths, Yuan Cao, and Karthik Narasimhan.
\newblock Tree of thoughts: Deliberate problem solving with large language models.
\newblock \emph{Advances in Neural Information Processing Systems}, 36, 2024.

\bibitem[Yu et~al.(2022)Yu, Jeong, Kim, Kim, and Chun]{orca}
Gyeong-In Yu, Joo~Seong Jeong, Geon-Woo Kim, Soojeong Kim, and Byung-Gon Chun.
\newblock Orca: A distributed serving system for $\{$Transformer-Based$\}$ generative models.
\newblock In \emph{16th USENIX Symposium on Operating Systems Design and Implementation (OSDI 22)}, pp.\  521--538, 2022.

\bibitem[Zheng et~al.(2023)Zheng, Yin, Xie, Huang, Sun, Yu, Cao, Kozyrakis, Stoica, Gonzalez, et~al.]{sglang}
Lianmin Zheng, Liangsheng Yin, Zhiqiang Xie, Jeff Huang, Chuyue Sun, Cody~Hao Yu, Shiyi Cao, Christos Kozyrakis, Ion Stoica, Joseph~E Gonzalez, et~al.
\newblock Efficiently programming large language models using sglang.
\newblock \emph{arXiv preprint arXiv:2312.07104}, 2023.

\bibitem[Zhou et~al.(2023)Zhou, Yan, Shlapentokh-Rothman, Wang, and Wang]{zhou2023language}
Andy Zhou, Kai Yan, Michal Shlapentokh-Rothman, Haohan Wang, and Yu-Xiong Wang.
\newblock Language agent tree search unifies reasoning acting and planning in language models.
\newblock \emph{arXiv preprint arXiv:2310.04406}, 2023.

\end{thebibliography}
